\documentclass{ecai} 

\usepackage{latexsym}
\usepackage{amssymb}
\usepackage{amsmath}
\usepackage{amsthm}
\usepackage{booktabs}
\usepackage{enumitem}
\usepackage{graphicx}
\usepackage{color}

\usepackage{booktabs}
\usepackage{multirow}
\usepackage{cleveref}

\newcommand{\BibTeX}{B\kern-.05em{\sc i\kern-.025em b}\kern-.08em\TeX}

\begin{document}


\begin{frontmatter}


\paperid{6678} 


\title{Learning to Reason: Temporal Saliency Distillation for Interpretable Knowledge Transfer.}


\author{\fnms{Nilushika Udayangani}~\snm{Hewa Dehigahawattage}\thanks{Corresponding Author. Email: hewadehigaha@student.unimelb.edu.au}}
\author{\fnms{Kishor}~\snm{Nandakishor}}
\author{\fnms{Marimuthu}~\snm{Palaniswami}} 

\address{Department of Electrical and Electronic Engineering, University of Melbourne Parkville VIC 3052, Australia}





\begin{abstract}
Knowledge distillation (KD) has proven effective for model compression by transferring knowledge from a larger network (teacher) to a smaller network (student). 
Current KD in time series is predominantly based on logit and feature aligning techniques originally developed for computer vision tasks. These methods do not explicitly account for temporal data and fall short in two key aspects. First, the mechanisms by which the transferred knowledge helps the student model's learning process remain unclear, due to uninterpretability of logits and features. 
Second, these methods transfer only limited knowledge, primarily replicating the teacher’s predictive accuracy. As a result, student models often produce predictive distributions that differ significantly from those of their teachers, hindering their safe substitution for teacher models.
In this work, we propose transferring interpretable knowledge by extending conventional logit transfer to convey not just the \textit{right prediction} but also the \textit{right reasoning} of the teacher.
Specifically, we induce other useful knowledge from the teacher logits, termed temporal saliency, which captures the importance of each input timestep to the teacher's prediction.
 By training the student with \textbf{T}emporal \textbf{S}aliency \textbf{D}istillation \textbf{(TSD)}, we encourage it to make predictions based on the same input features as the teacher. 
TSD requires no additional parameters or architecture-specific assumptions.
We demonstrate that TSD effectively improves the performance of baseline methods while also achieving desirable properties beyond predictive accuracy. We hope our work establishes a new paradigm for interpretable knowledge distillation in time series analysis.

\end{abstract}

\end{frontmatter}


\section{Introduction}

Time series classification~(TSC) has become fundamental to modern decision-making systems, enabling critical applications from patient health monitoring to industrial fault detection and environmental sensing. While deep neural networks~(DNN) achieve exceptional classification performance with large parameter spaces, deploying them on resource-constrained edge devices is hindered by high computational demands. This necessitates smaller, more efficient networks, which, however, often lack the inductive biases to learn representations from training data alone.
Knowledge distillation (KD) has proven that a smaller model (student) can still be trained to match the performance of a larger model (teacher), by learning to match the teacher’s logit distribution~\cite{ba_deep_2014,hinton_distilling_2014}. 
This allows efficient models that require only a fraction of the teacher’s computational resources, without sacrificing predictive accuracy. 
Later works extend KD to match intermediate features, beyond final layer logits~\cite{romero_fitnets_2014, park_relational_2019, zagoruyko_paying_2017, ahn_variational_2019}.
However, current KD literature largely focuses on improving student generalization (i.e., matching the teacher’s accuracy), which does not necessarily guarantee the transfer of well-established properties of the teacher model, such as robustness~\cite{stanton2021does}. This is because similar accuracy can often be achieved by classifiers with qualitatively different decision boundaries. 
As a result, KD may fail to live up to its conventional understanding, often leading to student models with predictive distributions that differ significantly from their teachers.

\begin{figure}[tb!]
\centering
\includegraphics[width=8cm]{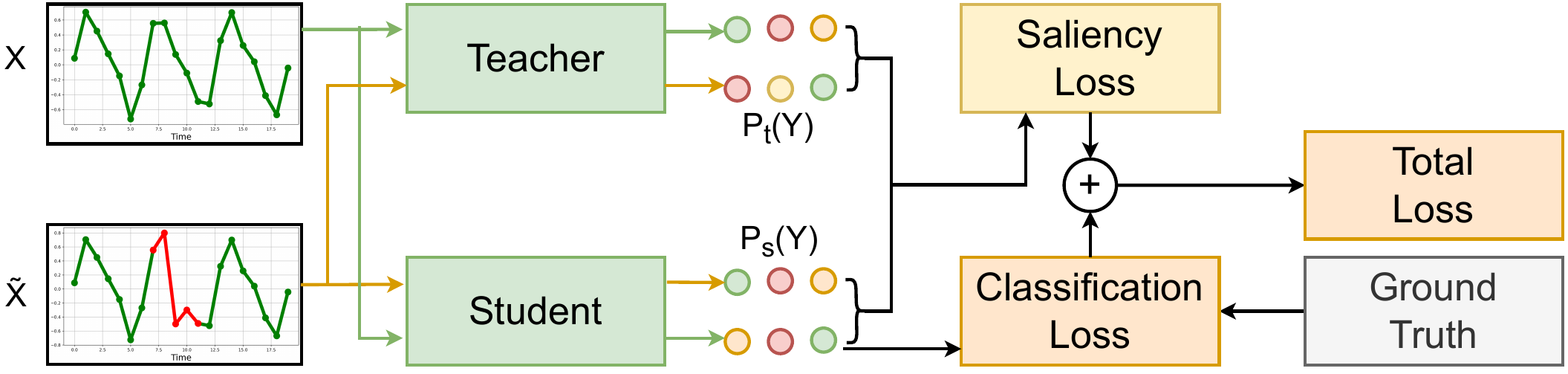}
\caption{\textbf{An overview of the proposed TSD}. The student network learns the target task by minimizing the classification loss while mimicking the temporal saliency observed by the teacher network.}
\label{pic:OverviewXL}
\end{figure}

\begin{figure*}[tb!]
    \centering
    \begin{minipage}[b]{0.28\textwidth}
        \centering
        \includegraphics[width=\textwidth]{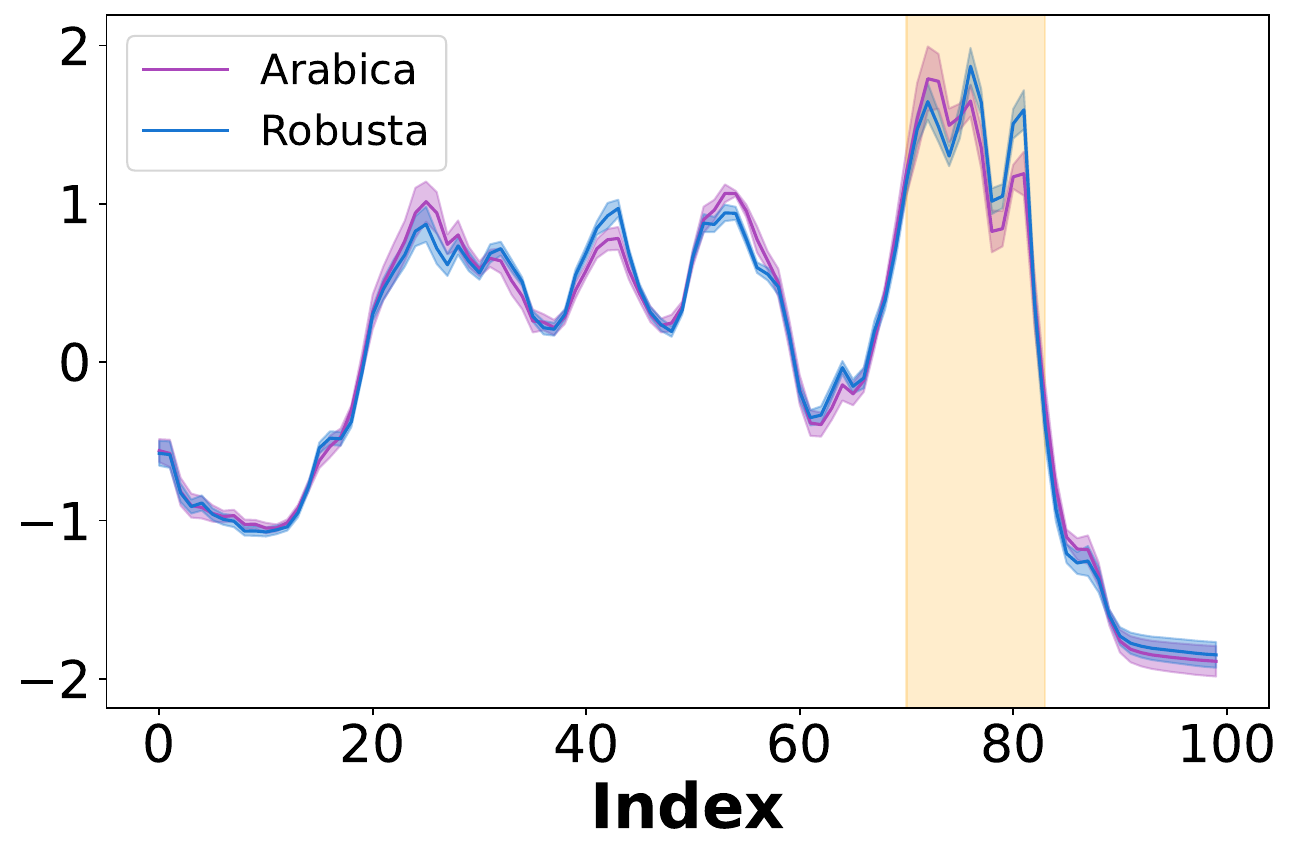}
    \end{minipage}
        \begin{minipage}[b]{0.3\textwidth}
        \centering
        \includegraphics[width=\textwidth]{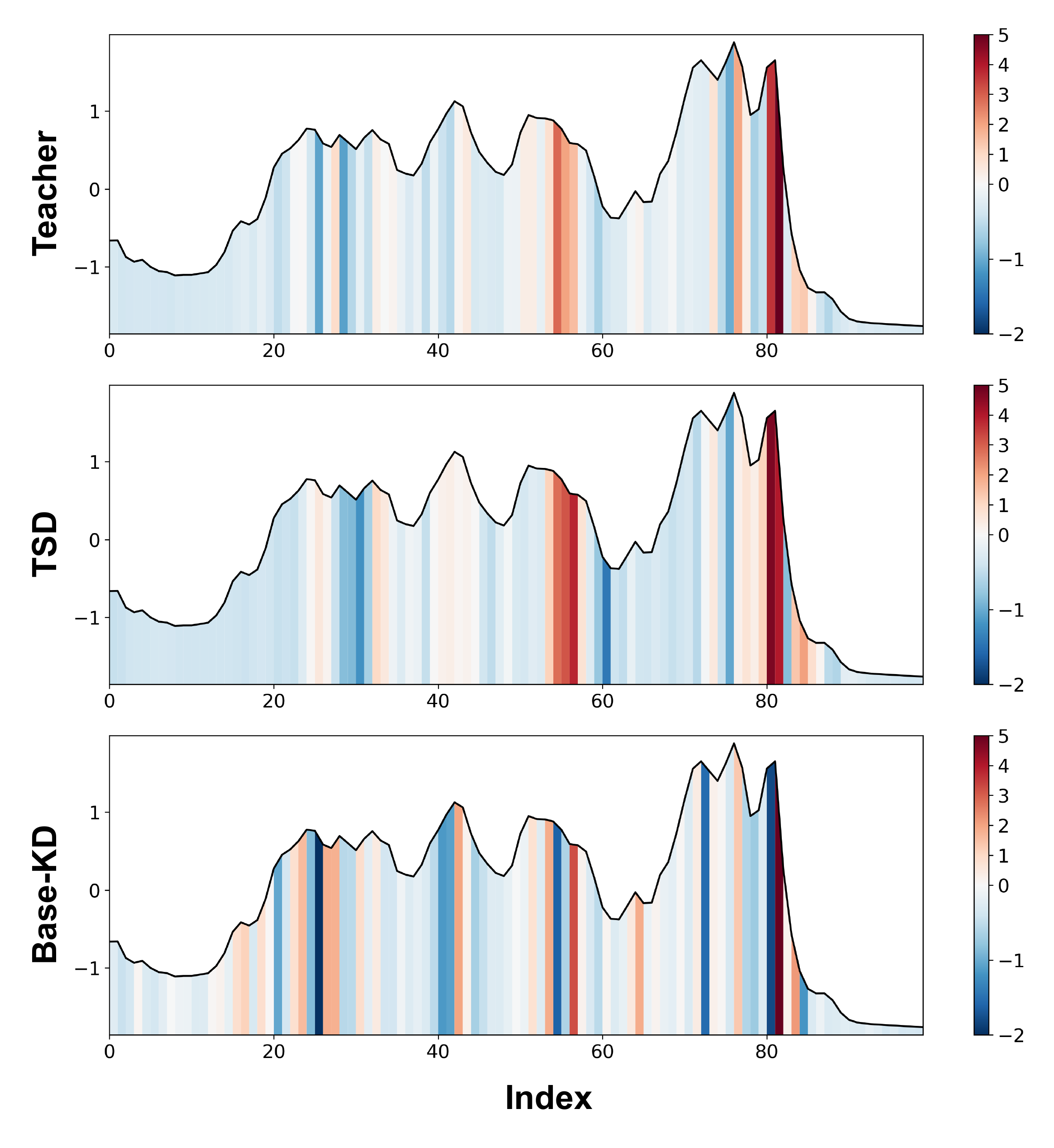}
    \end{minipage}
         \begin{minipage}[b]{0.3\textwidth}
        \centering
        \includegraphics[width=\textwidth]{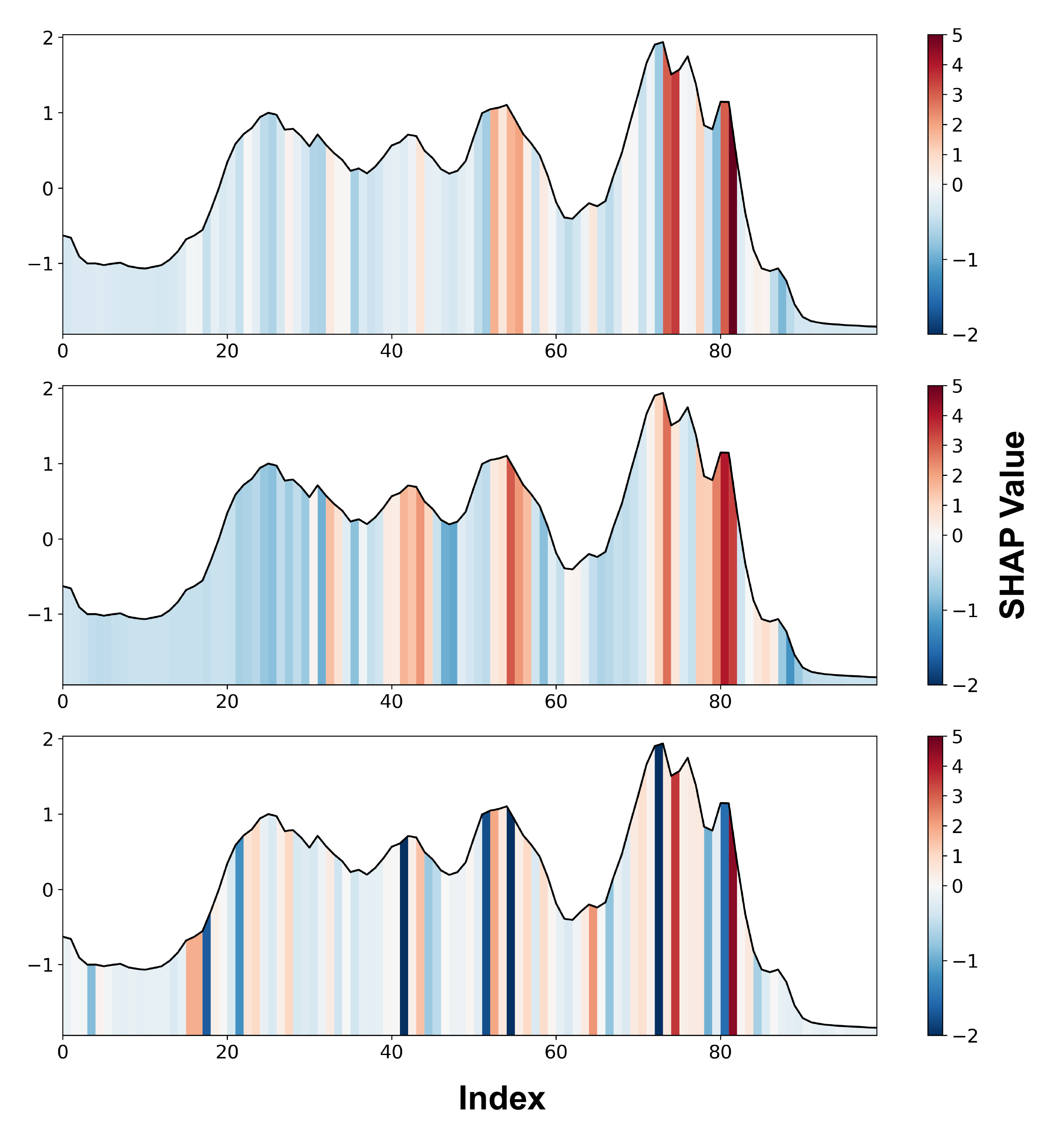}
    \end{minipage}
   \caption{\textbf{TSD transfers the \textit{right reasoning}}. \textbf{(Left)} Mean signals for the Coffee dataset for each class, the width representing the standard deviation. Compared to Robusta, Arabica beans have lower caffeine and chlorogenic acid content, contributing to their finer taste. The shaded area provides information about the caffeine content, which exhibits class-discriminative power compared to other regions~\cite{briandet1996discrimination,delaney2021instance}. \textbf{(Center)} Contribution of each input index to the class prediction for Robusta using SHAP analysis~\cite{lindberg2017unified}. Red indicates a positive contribution, and blue indicates a negative one. The TSD student exhibits input feature contributions consistent with the teacher, even in regions that are less class-discriminative. \textbf{(Right)} Same visualization for Arabica. }
    \label{fig:into_figs}
\end{figure*}

\begin{figure*}[tb!]
\centering
    \begin{minipage}[b]{0.26\textwidth}
        \centering
        \includegraphics[width=\textwidth]{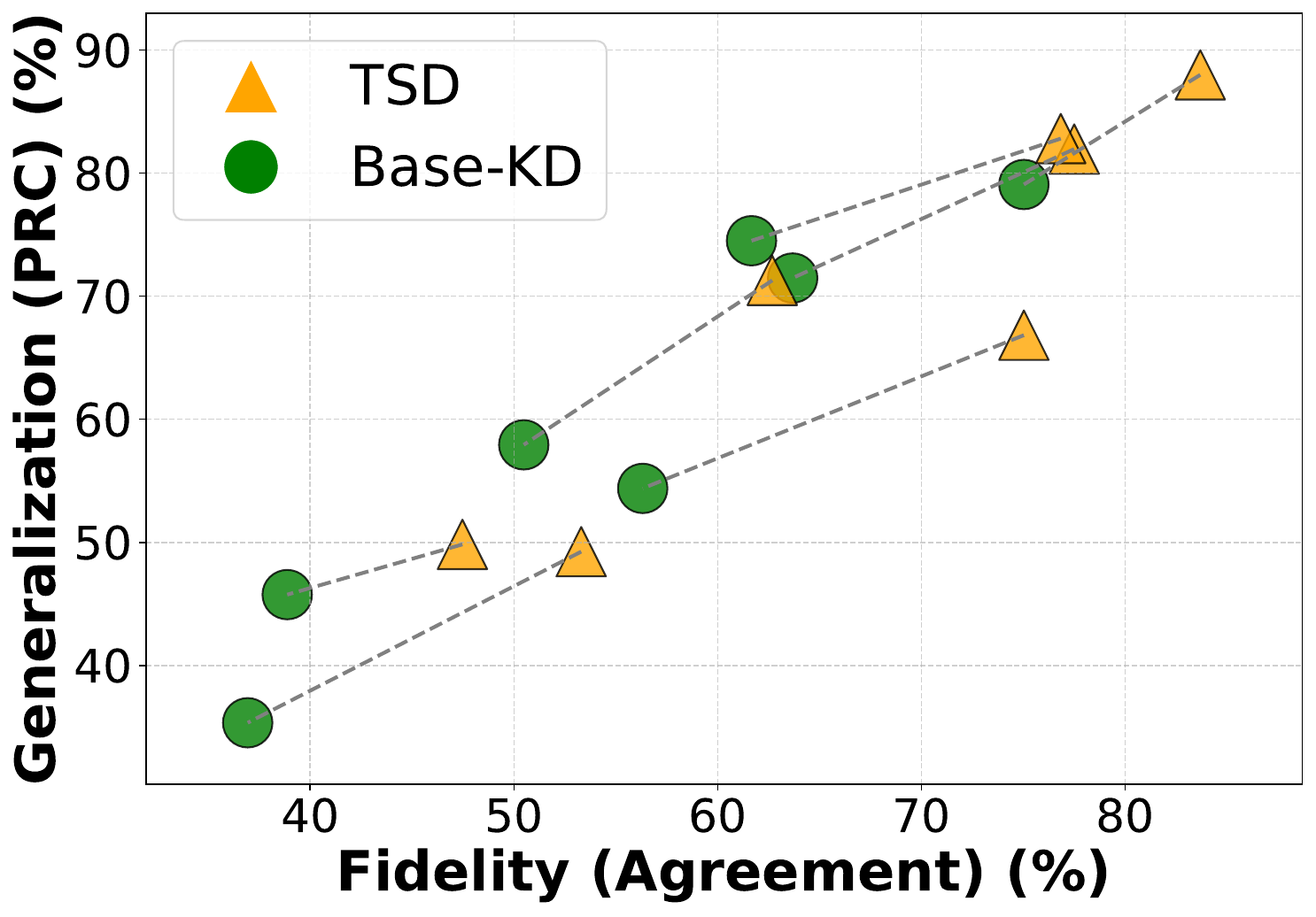}
    \end{minipage}
    \begin{minipage}[b]{0.278\textwidth}
        \centering
        \includegraphics[width=\textwidth]{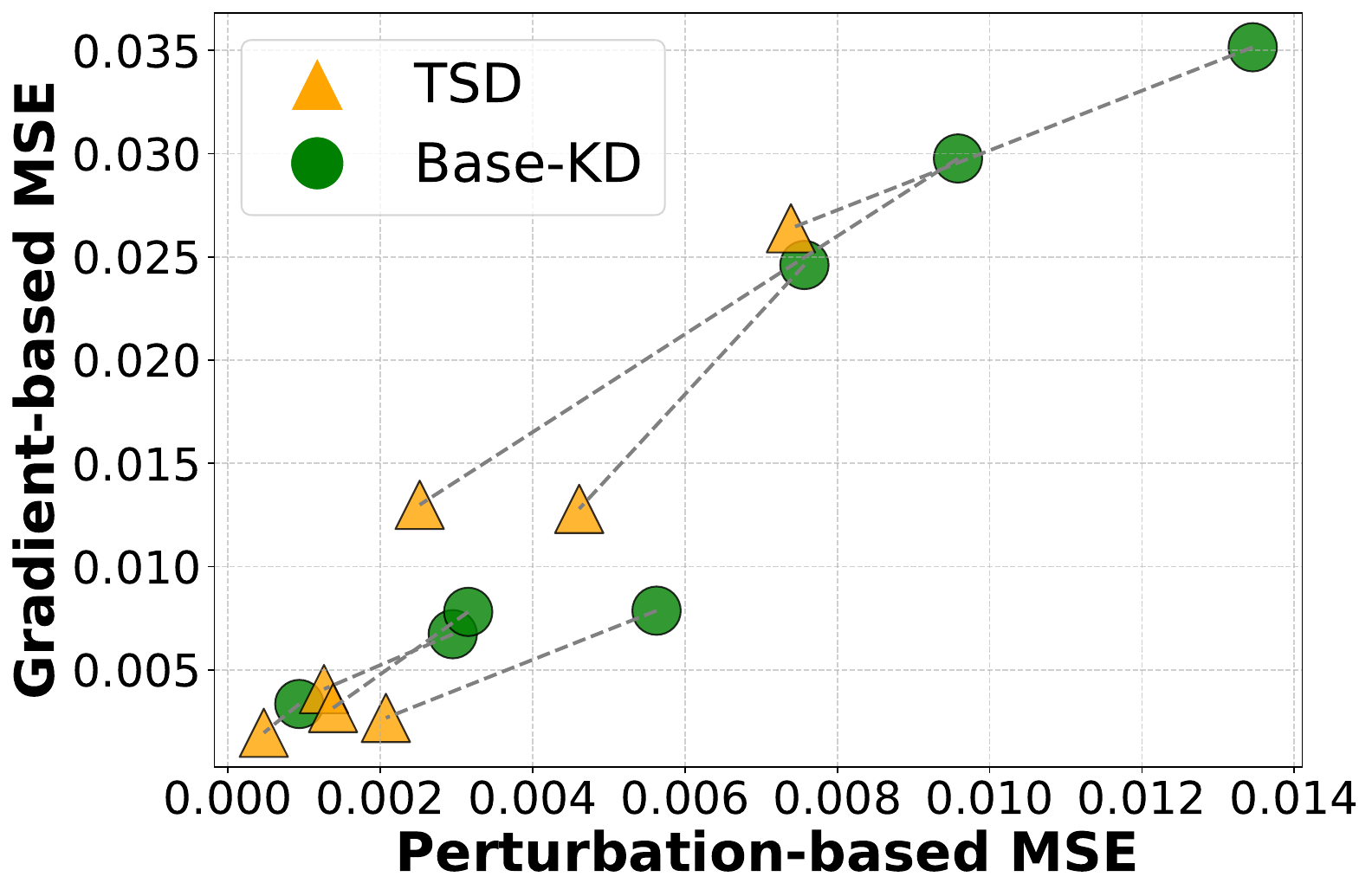}
    \end{minipage}
   \caption{\textbf{TSD transfer properties beyond generalization}. \textbf{(Left)} TSD enables student models to achieve both good generalization and good fidelity. Generalization is measured by AUC-PRC, and fidelity by test agreement on seven multiclass UCR datasets. \textbf{(Right)} TSD transfers the interpretability of the teacher model. The discrepancy between teacher and student saliency maps is measured using MSE.  Similarity improves for both gradient-based and perturbation-based saliency maps.}
    \label{fig:intro_figs2}
\end{figure*}

This motivates us to rethink the problem of knowledge extraction from time series models in a way that transfers properties beyond classification accuracy.
While current KD methods in time series primarily rely on direct logit matching adapted from computer vision tasks, we propose to extend logit matching to convey not just the \textit{right prediction} but also the \textit{right reasoning} from the teacher model.
In TSC, specific time steps featuring peaks, troughs, sudden changes, and patterns can be highly indicative of certain classes. For instance, in near-infrared spectrographs used to differentiate between Arabica and Robusta coffee, the region related to caffeine content is particularly discriminative (see \Cref{fig:into_figs}). 
By \textit{right reasoning}, we mean transferring knowledge about these salient input features, revealing the underlying rationale behind a teacher prediction.
But how can we encapsulate the teacher logits to include the \textit{right reasoning} at the same time? To solve this, we draw inspiration from saliency methods for explainable time series~\cite{tonekaboni2020went,parvatharaju_learning_2021}. 
Specifically, we induce a temporal saliency for each time step from teacher logits based on its contribution to the prediction. We quantify temporal saliency by measuring the sensitivity of the predictive distribution, comparing it with a perturbed version where only the target time step is altered. 
We propose training the student model to match the temporal saliency for each time step, as perceived by the teacher model, which we term Temporal Saliency Distillation (TSD). The overview of TSD is presented in~\Cref{pic:OverviewXL}.

We recognize the following benefits of TSD over conventional KD methods. TSD transfers interpretable knowledge, in contrast to direct logit/feature transfer methods, where it is hard to explain how the transferred knowledge benefits the student.
By encouraging the student to focus on the same input features as the teacher, TSD enables the student to maintain the same degree of model interpretability as the teacher.
\Cref{fig:into_figs} visualizes how students trained with TSD observe consistent temporal saliency with the teacher, even in regions other than class-discriminative ones, and achieve consistent explanations with the teacher (\Cref{fig:intro_figs2}).
Recent findings~\cite{stanton2021does} decouple the understanding of KD into generalization: the performance of a student in predicting unseen, in-distribution data, and fidelity: the ability of a student to match the teacher's predictions, showing that achieving good fidelity is vital but extremely difficult. We observe that transferring temporal saliency improves the similarity between teacher--student predictive distributions, leading to student models with both good generalization and fidelity (see \Cref{fig:intro_figs2}).
TSD integrates the \textit{right reasoning} directly into predictive distributions, without pre-computed explanations~\cite{alharbi2021learning,parchami2024good} or assumptions about network-specific layer properties~\cite{guo2023class}. Hence, TSD remains model-agnostic, enabling its application across diverse model architectures.
In short, we make following contributions:
 \begin{itemize}
     \item We propose a novel KD framework for time series model compression, extending logit transfer to incorporate not only the \textit{right prediction} but also the \textit{right reasoning} of the teacher. Our KD features a novel loss function that captures interpretable knowledge through perturbation-based temporal saliency. 
     \item Without requiring additional network modules or multi-layer feature extraction, we show TSD remains model agnostic and robust across various time series classification architectures, supporting both similar and disparate teacher-student configurations.
     \item We show that TSD achieves superior performance compared to state-of-the-art KD approaches, while also exhibiting desirable properties beyond generalization.
 \end{itemize}


\section{Background}
\paragraph{Saliency Methods.}The purpose of saliency methods is to highlight input features that are significant for a model's prediction. They can be categorized based on how they interact with the model to produce importance scores: 1) gradient-based methods use the gradients of the predictions with respect to input features, 2) perturbation-based methods assess the impact of altering input features on model predictions, and 3) attention-based methods use attention layers to generate importance scores. While originally developed for image classification, adapting these methods to time series requires careful consideration~\cite{crabbe2021explaining}.
When applied to time series, gradient-based methods tend to capture local variations of features, often disregarding time ordering~\cite{crabbe2021explaining}. Attention-based methods depend on specific architectures built around the attention mechanism~\cite{vaswani2017attention}, making them less suitable for model-agnostic use cases.
This work focuses on perturbation-based saliency due to its straightforward and intuitive reflection of how altering specific time steps affects model behavior.

\paragraph{Knowledge Distillation.}KD, introduced by \cite{bucilua2006model,hinton_distilling_2014}, demonstrates that smaller models can achieve comparable or superior performance through knowledge transfer from larger models. This process involves matching the teacher's softened logits with those of the student, adjusted by a temperature hyper-parameter to amplify the contribution of negative logits. Incorporating intermediate feature representations alongside final-layer logits has further improved performance, establishing state-of-the-art results~\cite{romero_fitnets_2014,zagoruyko_paying_2017,ahn_variational_2019,park_relational_2019}. However the mechanism by which this ``dark knowledge'' aids the student model remains unclear, limiting understanding and improvements in KD. Additionally, these methods transfer very limited knowledge from logits and features, focusing solely on predictive accuracy~\cite{stanton2021does}. 
To address these limitations, few recent studies~\cite{guo2023class,alharbi2021learning,parchami2024good} propose the transfer of explanations, grounded in explainable AI, which have emerged to enhance transparency in black-box DNN. Explainability transferring seek to convey class discriminative features to the student model through various means. 
Guo et al. \cite{guo2023class} proposed using CAM~\cite{zhou2016learning}, while other works~\cite{zeyu2023grad,alharbi2021learning,parchami2024good} align explanations generated by post-hoc methods such as GradCAM~\cite{selvaraju2017grad} and SHAP~\cite{lindberg2017unified}.
Explanation-based knowledge advances traditional logit- and feature-based knowledge, by enabling the transfer of properties beyond predictive accuracy, such as consistent explainability~\cite{alharbi2021learning,parchami2024good}. The mechanism behind how they improve the student model is transparent, as they guide the student on which input features to focus on, transferring the ability to identify the class-discriminative regions~\cite{guo2023class}.

However, the aforementioned works either rely on pre-calculated explanations generated by post-hoc explainers~\cite{alharbi2021learning}, which add extra computation, or require modifications to the model's structure to produce explanations, assuming access to model weights, which may not always be feasible due to privacy concerns~\cite{guo2023class}. Further, originating from computer vision tasks, these methods typically overlook time-related correlations when producing explanations~\cite{crabbe2021explaining} or assume properties of network-specific components (e.g., convolutional layers) that do not have clear analogies in time series models, like recurrent networks~\cite{guo2023class,alharbi2021learning}. While KD remains understudied in time series analysis, none of the existing works explore how explanation-based knowledge can be extracted and transferred in time series model compression. In this paper, we bridge this gap by proposing a novel KD framework for TSC that offers both high interpretability and competitive performance.

\section{Our Method}

\subsection{Knowledge Distillation for Time Series Classification}
A time series can be represented as \begin{math}
\boldsymbol{X} = \left\{\boldsymbol{x}^{(1)}, \boldsymbol{x}^{(2)},\cdots, \boldsymbol{x}^{(T)}\right\}
\end{math}, where $\boldsymbol{x}^{(i)} \in \mathbb{R}^n$, $i=1,\cdots,T$ and $\boldsymbol{X}\in \mathbb{R}^{n\times T}$.
A time series dataset $\mathcal{D}$ contains M pairs $(\boldsymbol{X}_i, \boldsymbol{Y}_i), i= 1, \cdots,M$, where $\boldsymbol{Y}_i \in \mathbb{R}^C$ represents corresponding one-hot encoded class label $c\in [1,C]$. 
TSC involves training a classifier $\mathcal{F}$, mapping input $\boldsymbol{X}_i$ to their corresponding label $\boldsymbol{Y}_i$. The goal of the KD is to train a student model (a shallow model) as $\mathcal{F}$ on our time series dataset $\mathcal{D}$ by leveraging the knowledge acquired by a pre-trained teacher model (a deep model). This is achieved by minimizing a distillation loss function that can be typically represented as:
\begin{equation}\label{eq_lkd}
L_{\mathrm{KD}} =\sum _{\boldsymbol{X}_i} \ell\left(f_t(\boldsymbol{X}_i),f_s(\boldsymbol{X}_i)\right)
\end{equation} where $f_t$ and $f_s$ represent functions for teacher and student models respectively and $\ell$ represents a similarity measure used to match the metrics of the teacher and student models.

\subsection{Temporal Saliency as Knowledge}
Identifying time steps or short subsequences which are maximally representative of a class is central to many traditional machine learning methods~\cite{ye_time_2009,patel_mining_2002,zhu_time_2018}.
Deep learning models naturally detect class-discriminative regions through architecture-specific mechanisms (CNNs via average activations~\cite{guo2023class}, RNNs via memory weights, transformers via attention), with recent studies in explainable artificial intelligence confirming that the ability to identify and prioritize these salient time steps is crucial for performance~\cite{delaney2021instance,parvatharaju_learning_2021,crabbe2021explaining}. 

In this work, we propose transferring the temporal saliency of individual time steps or regions perceived by the teacher model to improve the representations learned by the student model. We formally define temporal saliency as assigning each time step, $\boldsymbol{x}^{(t)}$, an importance score that quantifies the contribution of that time step to the output class distribution, $P(\boldsymbol{Y}\mid \boldsymbol{X})$. Building on perturbation-based explainability~\cite{fong2019understanding,parvatharaju_learning_2021}, we propose to define importance through the predictive distributional shift under KL-divergence, comparing original predictions to those from perturbed input sequences.

For a given input time series $\boldsymbol{X}$, let $ \mathbf{X}_{t: (t+z)} \in \mathbb{R}^{n \times z}$ be a subsequence, $\left\{\boldsymbol{x}^{(t)}, \boldsymbol{x}^{(t+1)},\cdots,\boldsymbol{x}^{(t+z-1)}\right\}$ including the time steps from $t$ to $t+z$ where $t,z\in [T-1], t+z\leq T$. Note that when $z=1$, $ \mathbf{X}_{t: (t+1)} = \boldsymbol{x}^{(t)} \in \mathbb{R}^{n}$ corresponds to a single time step.
Our goal is to assign a temporal saliency score $\mathbf{S}\left(t, z\right)$ to the region of contigous timesteps $\mathbf{X}_{t: (t+z)} $ (or single time step $\boldsymbol{x}^{(t)}$).

Let $\tilde{\boldsymbol{X}}(t,z)$ be a perturbed version of $\boldsymbol{X}$ that obtained via perturbing the time steps in $\mathbf{X}_{t: (t+z)} $. To generate $\tilde{\boldsymbol{X}}(t,z)$, we need to choose new values for replacement. Following previous works~\cite{parvatharaju_learning_2021,guilleme2019agnostic}, we choose replacements to be corresponding time steps from an opposing class instance $\mathbf{X^o}$, selected from the background dataset: $ \tilde{\boldsymbol{X}}(t,z) = \mathbf{X}_{1: t};\mathbf{X^o}_{t: (t+z)};\mathbf{X}_{(t+z): T+1} $. 
This approach prevents the systematic patterns in the time series from being dramatically altered and avoids creating perturbations unlike any the classifier has encountered~\cite{parvatharaju_learning_2021}.

We define the $\mathbf{S}\left(t, z\right)$ to be the importance of the time steps in $\mathbf{X}_{t: (t+z)} $ for the model to issue the prediction, quantified by the shift in predictive distribution when time steps in $\mathbf{X}_{t: (t+z)} $ is perturbed:
\begin{equation}\label{temp_sal}
\mathbf{S}\left(t,z\right)=\mathrm{KL}\left(P_\tau\left(Y \mid \boldsymbol{X}\right) \| P_\tau\left(Y \mid \tilde{\boldsymbol{X}}(t,z)\right)\right)
\end{equation}
where $P_\tau\left(Y \mid .\right)$ represents the probability distribution softened by temperature $\tau > 0 $ and is defined as:
\begin{equation}
P_\tau(Y \mid \boldsymbol{X})=\operatorname{softmax}\left(\frac{\operatorname{logits}(\boldsymbol{X})}{\tau}\right)
\end{equation}
The softmax temperature $\tau$ preserves the the primary information about the predicted class while $\tau>1$ enhances the contribution of other logits, increasing the relative importance of all classes.
We hypothesize that this softening helps capture more detailed information about how the entire distribution responds to a given perturbation, which we validate in~\cref{T_effect}.

We propose to align temporal saliency for a given $\mathbf{X}_{t: (t+z)} $, as defined in~\Cref{temp_sal}, among teacher and student. i.e. when the teacher considers certain time steps or regions important for predictions, the student should learn to assign similar importance to those same temporal locations. Therefore we define $f_t$ and $f_s$ in~\Cref{eq_lkd} as follows:
 \begin{equation}\label{eq_ft}
f_t = \left\{\frac{\mathbf{S}_t\left(t,z\right)}{ {\mu}_t}\right\}t,z\in [T-1], t+z\leq T
\end{equation} 
\begin{equation}\label{eq_fs}
f_s = \left\{\frac{\mathbf{S}_s\left(t,z\right)}{ {\mu}_s}\right\}t,z\in [T-1], t+z\leq T
\end{equation} 
where $\boldsymbol{S}_t$ and $\boldsymbol{S}_s$ represent temporal saliency perceived by teacher and student models respectively. To mitigate any significant scale differences between the teacher's and student's saliency measures and thereby ensure the stability of the loss function, we normalize each difference by the per-instance mean temporal saliency value, ${\mu}$ of respective models.
 As for the metric $\ell$ to measure the difference between $f_t$ and $f_s$, we utilize \textit{Smooth L1 loss}:
\begin{equation}\label{eq_salkd}
L_{\mathrm{TSKD}} = \sum_{\boldsymbol{X}_i} \ell_{\mathrm{smoothL1}}\left(f_t(\boldsymbol{X}_i), f_s(\boldsymbol{X}_i)\right)
\end{equation}
where $L_{\mathrm{TSKD}}$ represents the proposed temporal-saliency distillation loss. Since TSD captures the relative importance of the entire series to predictions, it transfers comprehensive knowledge by communicating both discriminative (associated with higher saliency values) and non-discriminative (linked to lower saliency values) regions. Additionaly, note that selections for $t$ defines the number of subsequences and $z$ defines the length of each subsequence in evaluating the temporal saliency, with both serving as hyperparameters of our loss function. We analyze these hyperparameters in the ablation study presented in~\cref{ablation_params}.


Note that TSD only accesses model outputs to compute the distillation loss and does not rely on architecture-specific details. 
As a result, it has the potential to operate seamlessly across different architectures without requiring modifications to model weights or additional regressors to align feature dimensions (see results in \cref{model_agnostic}).
\section{Experiments}
\subsection{Datasets and Implementation Details}
\paragraph{Teacher and Student Models.}
In our experiments, we use a ResNet~\cite{wang2017time} (a network primarily composed of convolutional layers), a Long Short-Term Memory~(LSTM) network~\cite{hochreiter1997long} (built upon recurrent blocks) and an InceptionTime network~\cite{ismail2020inceptiontime} (which is among the current state-of-the-art for TSC) as our teacher models. Smaller variations of ResNet, LSTM, InceptionTime and Fully Convolutional Networks (FCN)~\cite{wang2017time} are used as student models under different compression levels, achieved by varying the number of layers and output dimensions. The total number of parameters, model sizes, and network configuration for all constructed models are summarized in \Cref{tab:network_config}.

\paragraph{Datasets.} 
We conducted our experiments on the UCR-2015 archive~\cite{chen_ucr_2015}. For the main results, we selected 28 datasets,, following similar selections in recent works~\cite{xing_2022_efficient,campos_2023_lightts}. The datasets were categorized by series length (`short' <150, `medium', 150-500 and `long' >500)~\cite{xing_2022_efficient}, with emphasis on multi-class problem~\cite{campos_2023_lightts}. The subset included 10 short, 8
medium, and 10 long datasets, with 7 challenging datasets having many output classes. All series were standardized to length 100 via linear interpolation, z-normalized, and evaluated with the original train/test split with 20\% validation. Similar performance improvements were observed for the remaining datasets, as reported in the supplementary material~\cite{hewa_dehigahawattage_2025_16938636}.

\begin{table}[tb!]
\centering
\caption{Configuration of teacher and student networks. The output dimension indicates the hidden size for the LSTM and the output dimension of the first convolutional layer for InceptionTime, ResNet and FCN~\cite{wang2017time}. }
\label{tab:network_config}
\begin{scriptsize}
\setlength{\tabcolsep}{1pt} 
\renewcommand{\arraystretch}{1} 
\begin{tabular}{@{}clcccc@{}}
\multicolumn{1}{l}{}     &       &  {Num. Layers} & { Output Dim.} &  {Total Param.} & {Model Size(MB)} \\ \toprule
\multirow{2}{*}{Teacher} & Inception55-32 & 55          & 32          & 978440       & 0.9361         \\
                         & LSTM3-100   & 3           & 100         & 812008       & 0.7744         \\
                         & Resnet32-64 & 32          & 64          & 2016008      & 1.9315         \\ \midrule
\multirow{7}{*}{Student} & Inception28-16 & 28          & 16          & 121864       & 0.117          \\
                         & Inception19-8  & 19          & 8           & 5368         & 0.0054         \\
                         & LSTM2-32    & 2           & 32          & 51976        & 0.0496         \\
                         & LSTM1-8     & 1           & 8           & 1480         & 0.0014         \\
                         & Resnet15-4  & 15          & 4           & 2328         & 0.0025         \\
                         & Resnet15-2  & 15          & 2           & 736          & 0.0009         \\
                         & FCN10-8      & 10          & 8           & 4808         & 0.0049         \\
                         & FCN10-4      & 10          & 4           & 1384         & 0.0015         \\
                         & FCN7-8      & 7           & 8           & 2120          & 0.0022         \\
                         & FCN7-4      & 7           & 4           & 680          & 0.0008         \\ \bottomrule 
\end{tabular}
\end{scriptsize}
\end{table}

\paragraph{Implementation details.} 
We select the best teacher model from five random initializations based on validation area under the precision-recall curve (AUC-PRC)~\cite{wang2017time}. Student models are trained using TSD and multiple state-of-the-art KD methods (Base-KD~\cite{hinton_distilling_2014}, FitNet~\cite{romero_fitnets_2014}, RKD~\cite{park_relational_2019}, Att~\cite{zagoruyko_paying_2017}, DKD~\cite{zhao_decoupled_2022}, DT2W~\cite{qiao_class-incremental_2023}), alongside baseline models (Base) trained without KD. 
All students involving KD are trained using a combination of the distillation loss and the classification loss (cross-entropy loss):
\begin{equation*}
    L_{\mathrm{train}} = \alpha\times L_{\mathrm{CE}} + \beta\times L_{\mathrm{KD}}
    \label{total_loss}
\end{equation*}
where $\alpha$ and $\beta$ decide the contribution of classification loss $L_{\mathrm{CE}}$ and distillation loss $L_{\mathrm{KD}}$ for the total train loss $L_{\mathrm{train}}$, respectively.
For all experiments, $\alpha$ is fixed at 1, while $\beta$ is optimized via grid search ${0.1, 0.5, 1, 10, 100, 200}$.
Models are implemented in PyTorch~\cite{paszke_2019_pytorch} using Adam optimizer, batch size 32, maximum 500 epochs with 50-epoch patience, and learning rate decay of 0.5 at epochs 25, 30, 35. Initial learning rates are 0.01 for FCN7-4/8, LSTM3-100, LSTM2-32 models and 0.1 for other models. All results are averaged over five runs with different random initializations.

\paragraph{Evaluation Metrics.} Model generalization was evaluated using area under the receiver operating characteristic curve (AUC-ROC), average AUC-PRC, and accuracy on the test set. A win/tie/loss calculation was employed, where a model `wins' on a dataset, if it achieves the highest AUC-PRC. We prioritized AUC-PRC over other metrics due to its robustness to class imbalance. Additionally, we calculated the average rank across datasets for each model. Since averaged values were used for the students, comparisons with teacher models also employed averaged values over five runs (instead of top-1). 
We adopt two metrics from~\cite{stanton2021does} to measure model fidelity: 1) the average agreement between the student’s and teacher’s top-1 predictions:
\begin{equation*}
\text{Average Top-1 Agreement} = \frac{1}{M} \sum_{i=1}^M \mathbb{1}\left(y_{i,t} = y_{i,s}\right),
\end{equation*}
and 2) the average KL divergence from the teacher’s predictive distribution to the student’s, which captures fidelity across the entire output distribution:
\begin{equation*}
\text { Predictive KL }=\frac{1}{M} \sum_{i=1}^M\mathrm{KL}\left(P_{\tau,t}\left(Y \mid \boldsymbol{X_i}\right) \| P_{\tau,s}\left(Y \mid {\boldsymbol{X_i}}\right)\right).
\end{equation*}

\begin{table*}[tb!]
\caption{Comparison of results with state-of-the-art methods. Experiments distill an LSTM3-100 teacher into an LSTM1-8 student. Reported values represent average AUC-PRC over five runs, except for the teacher model, where both top-1 and average results are shown (with the latter used for comparisons). \textbf{Bold} values indicate the best AUC-PRC among student models.}
\label{tab:sota}
\centering
\begin{scriptsize}
\setlength{\tabcolsep}{6pt} 
\renewcommand{\arraystretch}{1.2} 
\begin{tabular}{@{}ccccccccccccc@{}}
Database & \begin{tabular}[c]{@{}c@{}}{Teach.}\\ (max)\end{tabular} & \begin{tabular}[c]{@{}c@{}}{Teach.}\\ (avg)\end{tabular} & Base & \begin{tabular}[c]{@{}c@{}}Base\\ -KD\end{tabular} & DKD & Att. & \begin{tabular}[c]{@{}c@{}}RKD\\ -D\end{tabular} & \begin{tabular}[c]{@{}c@{}}RKD\\ -A\end{tabular} & \begin{tabular}[c]{@{}c@{}}RKD\\ -DA\end{tabular} & {Fitnet} & {DT2W} & Ours \\ \midrule
Computers & 62.46 & 59.58 & 57.16 & 64 & 63.78 & 63.72 & 62.66 & 64.12 & 62.76 & 58.9 & 62.17 & \textbf{64.74} \\
UWaveGesture.All & 94.75 & 91.72 & 77.15 & 79.07 & 80.58 & 79.31 & 78.34 & 78.99 & 77.51 & 81.96 & 81.35 & \textbf{87.98} \\
Strawberry & 94.25 & 87.91 & 70.68 & 73.06 & 71.13 & 89.09 & 90.05 & 91.53 & \textbf{92.08} & 87.1 & 59.54 & 89.67 \\
BeetleFly & 97.2 & 81.64 & 81.17 & 86.15 & 84.16 & 88.06 & 90.98 & \textbf{93.57} & 90.03 & 79.78 & 76.99 & 90.27 \\
wafer & 99.34 & 81.19 & 98.68 & 98.55 & 98.83 & \textbf{98.98} & 98.7 & 98.81 & 98.72 & 98.95 & 89.35 & 98.9 \\
CBF & 99.69 & 72.44 & 69.1 & 75.23 & 90.07 & 76.6 & 89.74 & 96.29 & 93 & 99.19 & 51.29 & \textbf{99.65} \\
Adiac & 70.61 & 65.79 & 37.5 & 45.75 & 46.36 & 49.21 & 41.93 & 43.3 & 39.94 & 37.67 & 43.66 & \textbf{49.83} \\
Lighting2 & 69.16 & 61.84 & 65.59 & 69.03 & 68.24 & 69.65 & 68.29 & \textbf{70.01} & 68.81 & 65.24 & 66.73 & 67.77 \\
ItalyPowerDemand & 99.24 & 98.21 & 94.22 & 98.63 & 97.68 & 97.55 & 98.37 & 99.07 & 98.78 & 97.23 & 97.94 & \textbf{99.2} \\
yoga & 75.88 & 73.04 & 57.97 & 67.41 & 66.06 & \textbf{70.61} & 68.36 & 67.78 & 69.33 & 60.1 & 52.12 & 68.24 \\
Trace & 73.84 & 54.39 & 64.76 & 69.25 & 68.21 & 61.61 & 72.71 & 72.94 & 72.48 & \textbf{74.11} & 53.14 & 73.5 \\
ShapesAll & 72.34 & 71.04 & 29.82 & 35.34 & 44.46 & 35.11 & 30.27 & 30.98 & 31.98 & 35.97 & 33.55 & \textbf{49.24} \\
Beef & 62.81 & 50.56 & 37.41 & 46.86 & 44.03 & 49.53 & 44.71 & 48.95 & 50.65 & 44.97 & 37.35 & \textbf{57.28} \\
Herring & 64.3 & 55.6 & 50.28 & 54.62 & 57.06 & \textbf{61.01} & 56.36 & 55.25 & 55.77 & 57.77 & 58 & 59.74 \\
MiddlePhalanx.Corr. & 66.62 & 62.03 & 60.37 & 65.49 & 67.18 & 65.54 & 66.48 & \textbf{68.21} & 66.79 & 63.24 & 63.17 & 63.47 \\
FordA & 97.18 & 65.66 & 90.33 & 93.96 & 93.2 & 95.58 & 96.34 & 96.29 & 96.12 & 86.9 & 49.73 & \textbf{97.36} \\
SwedishLeaf & 92.13 & 89.65 & 70.75 & 71.47 & 72.95 & 76.39 & 70.87 & 72.94 & 71.11 & 71.72 & 73.14 & \textbf{81.95} \\
FaceAll & 87.16 & 82.77 & 46.22 & 54.38 & 59.82 & 58.21 & 52.68 & 54.6 & 57.88 & 56.08 & 55.92 & \textbf{66.83} \\
StarLightCurves & 97.69 & 96.8 & 93.11 & 96.09 & 95.59 & 95.54 & 95.08 & 93.63 & 94.37 & 91.75 & 92.19 & \textbf{96.91} \\
ECG200 & 79.35 & 69.84 & 67.55 & 68.39 & 68.5 & 70.35 & 73.09 & \textbf{73.79} & 73.07 & 71.85 & 66.1 & 71.49 \\
ECGFiveDays & 93.14 & 83.32 & 81.67 & 82.28 & 83.33 & \textbf{91.57} & 85.32 & 89.02 & 85.96 & 91.45 & 76.7 & 88.87 \\
OliveOil & 55.02 & 36.15 & 25.37 & 25.32 & 25.06 & 32 & 34.38 & 37.76 & 35 & 44.67 & 37.93 & \textbf{46.92} \\
MoteStrain & 83.89 & 80.75 & 82.74 & 81.87 & 81.39 & 82.58 & 83.02 & 83.53 & 82.76 & 78.16 & 84.21 & \textbf{87.99} \\
SonyAIBORobotSurf. & 72.42 & 67.2 & 62.96 & 63.57 & 63.23 & 65.55 & 66.74 & \textbf{70.76} & 69.49 & 68.14 & 56.35 & 70.22 \\
SonyAIBORobotSur2. & 80.38 & 71.96 & 70.02 & 72.06 & 72.78 & 73.48 & 72.6 & 75.61 & 75.15 & 72.47 & 72.85 & \textbf{79.77} \\
Ham & 71.93 & 69.17 & 62.17 & 65.78 & 65.5 & \textbf{70.4} & 68.16 & 69.34 & 68.65 & 64.13 & 57.06 & 69.49 \\
NonInv.FetalECG1 & 86.11 & 82.29 & 58.08 & 57.92 & 56.96 & 64.21 & 61.02 & 58.4 & 55.09 & 65.4 & 55.01 & \textbf{71.28} \\
NonInv.FetalECG2 & 91.71 & 86.7 & 71.2 & 74.5 & 66.69 & 73.26 & 73.25 & 73.51 & 73.41 & 71.13 & 72.64 & \textbf{82.8} \\
 &  &  &  &  &  &  &  &  &  &  &  &  \\ \midrule
Avg. AUC-PRC & 81.81 & 73.19 & 65.5 & 69.14 & 69.74 & 71.6 & 71.09 & 72.46 & 71.67 & 70.57 & 63.44 & \textbf{76.12} \\
Avg. AUC-ROC & 87.58 & 80.66 & 77.00 & 80.36 & 80.83 & 82.04 & 82.82 & 83.65 & 82.78 & 80.23 & 75.30 & \textbf{85.04} \\
Avg. Acc. & 75.92 & 69.58 & 62.94 & 67.23 & 67.55 & 69.25 & 68.80 & 70.19 & 69.25 & 66.85 & 62.12 & \textbf{70.84} \\
Wins & - & - & 0 & 0 & 0 & 5 & 0 & 5 & 1 & 1 & 0 & \textbf{16} \\
Lose & - & - & 28 & 28 & 28 & 23 & 28 & 23 & 27 & 27 & 28 & \textbf{12} \\
Avg.Rank & - & - & 8.89 & 6.18 & 5.96 & 4.21 & 5.46 & 3.75 & 4.79 & 5.96 & 7.61 & \textbf{2.18} \\
Wins(with teacher) & - & 8 & 0 & 0 & 0 & 4 & 0 & 5 & 1 & 1 & 0 & \textbf{9} \\
Lose(with teacher) & - & 20 & 28 & 28 & 28 & 24 & 28 & 23 & 27 & 27 & 28 & \textbf{19} \\
Avg.Rank\tiny{(with teacher)} & - & 5.71 & 9.71 & 6.79 & 6.54 & 4.68 & 5.93 & 4.14 & 5.18 & 6.54 & 8.32 & \textbf{2.46} \\ \bottomrule
\end{tabular}
\end{scriptsize}
\end{table*}

\begin{figure*}[tb!]
    \centering
    \begin{minipage}[b]{0.33\textwidth}
        \centering
        \includegraphics[width=\textwidth]{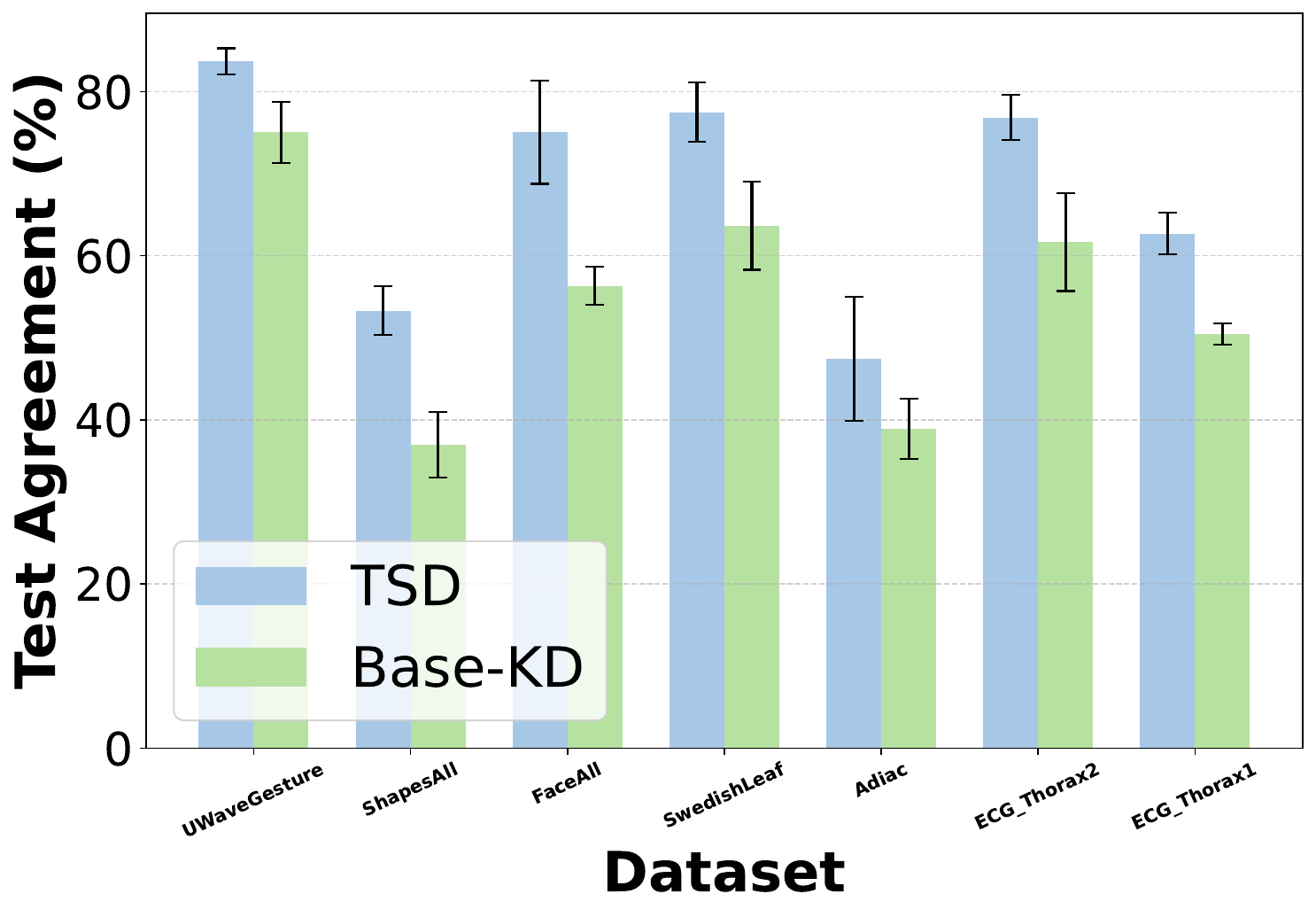}
    \end{minipage}
    \hfill
    \begin{minipage}[b]{0.33\textwidth}
        \centering
        \includegraphics[width=\textwidth]{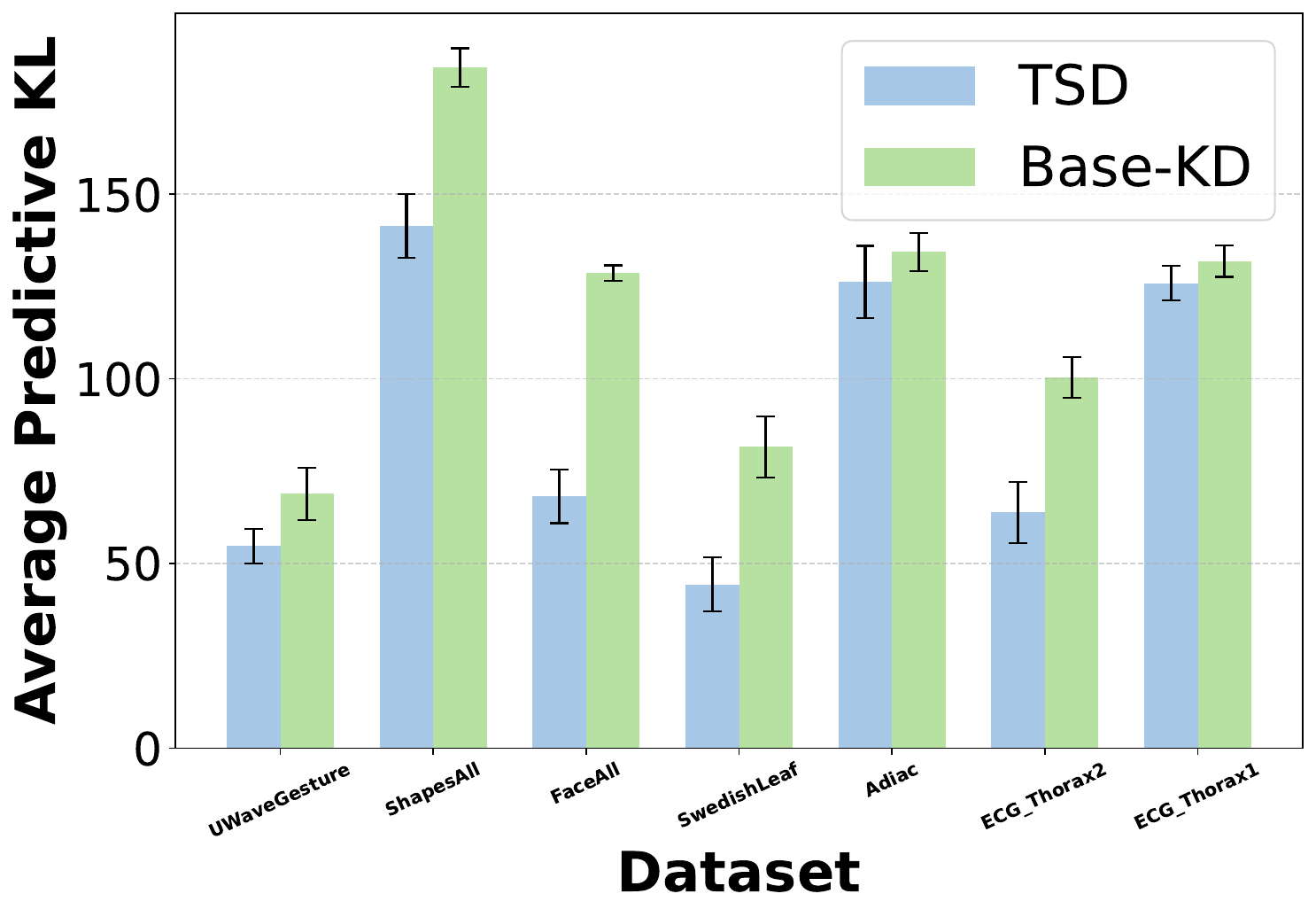}
    \end{minipage}
    \hfill
    \begin{minipage}[b]{0.33\textwidth}
        \centering
        \includegraphics[width=\textwidth]{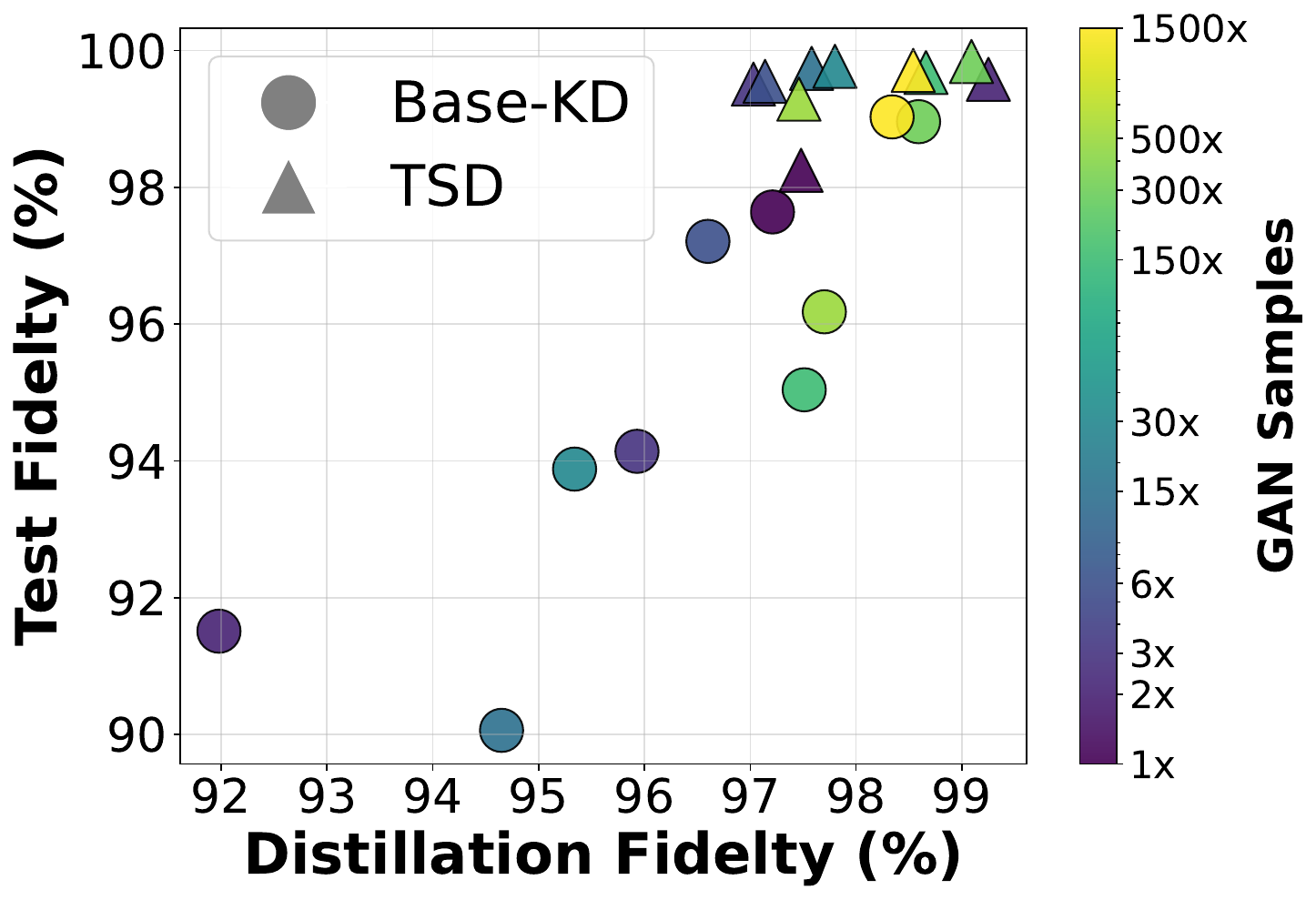}
    \end{minipage}
    \caption{Evaluation of test fidelity across seven multi-class UCR datasets: \textbf{(left)} average top-1 agreement; \textbf{(center)} average predictive KL divergence. \textbf{(Right)} Train-test fidelity across various expansion ratios of the distillation set with GAN generated synthetic data.}
    \label{fig:fidelty_figsAll}
\end{figure*}

\begin{table*}[tb!]
\caption{Comparison with baselines using teacher and student from the same model families. Varying compression factors are achieved, as
indicated.}
\label{tab:sim_arch}
\begin{scriptsize}
\setlength{\tabcolsep}{2.9pt}
\renewcommand{\arraystretch}{1.5}

\begin{tabular}{c|ccc|ccc|ccc|ccc|ccc|ccc}
Teacher      & \multicolumn{3}{c|}{LSTM3-100}   & \multicolumn{3}{c|}{LSTM3-100}    & \multicolumn{3}{c|}{Resnet32-64}  & \multicolumn{3}{c|}{Resnet32-64}  & \multicolumn{3}{c|}{Inception55-32} & \multicolumn{3}{c}{Inception55-32} \\ 
Student      & \multicolumn{3}{c|}{LSTM1-8}     & \multicolumn{3}{c|}{LSTM2-32}     & \multicolumn{3}{c|}{Resnet15-2}   & \multicolumn{3}{c|}{Resnet15-4}   & \multicolumn{3}{c|}{Inception19-8}               & \multicolumn{3}{c}{Inception28-16} \\
Compression  & \multicolumn{3}{c|}{550x}        & \multicolumn{3}{c|}{16x}          & \multicolumn{3}{c|}{2739x}        & \multicolumn{3}{c|}{866x}         & \multicolumn{3}{c|}{180x}           & \multicolumn{3}{c}{18x}            \\
             & Base & Base-KD & Ours           & Base  & Base-KD & Ours           & Base  & Base-KD & Ours           & Base  & Base-KD & Ours           & Base   & Base-KD  & Ours           & Base   & Base-KD  & Ours           \\
Avg.AUC-PRC & 65.5 & 69.14   & \textbf{76.12} & 70.18 & 74.46   & \textbf{80.84} & 71.05 & 72.36   & \textbf{74.98} & 81.11 & 81.83   & \textbf{84.17} & 59.83  & 60.05    & \textbf{63.48} & 78.88  & 79.65    & \textbf{81.95} \\
Wins         & 0    & 2       & \textbf{26}    & 1     & 3       & \textbf{24}    & 1     & 9       & \textbf{18}    & 4     & 5       & \textbf{20}    & 3      & 4        & \textbf{22}    & 3      & 7        & \textbf{18}    \\
Lose         & 28   & 26      & \textbf{2}     & 27    & 25      & \textbf{4}     & 27    & 19      & \textbf{10}    & 24    & 23      & \textbf{8}     & 25     & 24       & \textbf{6}     & 25     & 21       & \textbf{10}    \\
Avg. Rank    & 2.86 & 2.07    & \textbf{1.07}  & 2.86  & 2       & \textbf{1.14}  & 2.68  & 1.89    & \textbf{1.43}  & 2.5   & 1.96    & \textbf{1.5}   & 2.36   & 2.18     & \textbf{1.43}  & 2.43   & 2.11     & \textbf{1.46}  \\ \bottomrule
\end{tabular}

\caption{Comparison with baselines using teacher and student from the different model families. Varying compression factors are achieved, as
indicated.}
\label{tab:diff_arch}

\begin{tabular}{c|ccc|ccc|ccc|ccc|ccc|ccc}
Teacher      & \multicolumn{3}{c|}{Resnet32-64}  & \multicolumn{3}{c|}{Resnet32-64}  & \multicolumn{3}{c|}{Resnet32-64}  & \multicolumn{3}{c|}{Resnet32-64}  & \multicolumn{3}{c|}{Inception55-32} & \multicolumn{3}{c}{Inception55-32} \\ 
Student      & \multicolumn{3}{c|}{FCN10-4}       & \multicolumn{3}{c|}{FCN10-8}       & \multicolumn{3}{c|}{FCN7-4}       & \multicolumn{3}{c|}{FCN7-8}       & \multicolumn{3}{c|}{Resnet15-2}     & \multicolumn{3}{c}{FCN7-8}         \\
Compression  & \multicolumn{3}{c|}{1457x}        & \multicolumn{3}{c|}{419x}         & \multicolumn{3}{c|}{2970x}        & \multicolumn{3}{c|}{950x}         & \multicolumn{3}{c|}{1330x}          & \multicolumn{3}{c}{460x}           \\
             & Base  & Base-KD & Ours           & Base  & Base-KD & Ours           & Base  & Base-KD & Ours           & Base  & Base-KD & Ours           & Base   & Base-KD  & Ours           & Base   & Base-KD  & Ours           \\
Avg.AUC-PRC & 78.37 & 78.91   & \textbf{80.59} & 86.76 & 87.07   & \textbf{88.08} & 74.45 & 74.62   & \textbf{78.23} & 82.24 & 82.86   & \textbf{85.39} & 71.05  & 73.78    & \textbf{77.3}  & 82.17  & 82.47    & \textbf{84.93} \\
Wins         & 5     & 7       & \textbf{17}    & 6     & 8       & \textbf{16}    & 2     & 7       & \textbf{19}    & 4     & 5       & \textbf{21}    & 0      & 5        & \textbf{23}    & 3      & 9        & \textbf{19}    \\
Lose         & 23    & 21      & \textbf{11}    & 22    & 20      & \textbf{12}    & 26    & 21      & \textbf{9}     & 24    & 23      & \textbf{7}     & 28     & 23       & \textbf{5}     & 25     & 19       & \textbf{9}     \\
Avg. Rank    & 2.39  & 2       & \textbf{1.57}  & 2.29  & 1.86    & \textbf{1.71}  & 2.46  & 2.18    & \textbf{1.36}  & 2.36  & 2.14    & \textbf{1.39}  & 2.61   & 2.14     & \textbf{1.21}  & 2.25   & 2.14     & \textbf{1.46}  \\ \bottomrule
\end{tabular} 

\end{scriptsize}
\end{table*}

\subsection{Evaluation of TSD}
\paragraph{TSD produces students with good generalization.}
Table~\ref{tab:sota} presents the evaluation results of different KD variants across 28 datasets, including win/tie/loss calculations with and without the teacher. TSD, consistently outperforms all other distillation objectives in about 60\% of the datasets. The average rank of TSD is 2.18, which is significantly lower than that of other methods, indicating that TSD delivers competitive performance across the remaining 40\% datasets as well. None of the other methods achieved more than 18\% wins or ranks close to 2. Additionally, our method shows at least a 3.7\% improvement over other methods in terms of average AUC-PRC. All KD methods report lower ranks compared to the Base model, indicating that they all benefit from KD. Furthermore, TSD achieves the lowest average rank even in the calculations that include the teacher model. 
It is the only method to exceed the teacher in terms of average AUC-PRC and number of wins. Notably, although TSD only utilizes logit-level information to extract interpretable knowledge, it outperforms feature distillation methods that require additional network modules and multiple-layer information.

\paragraph{TSD is model agnostic.}\label{model_agnostic}
We evaluate effectiveness of TSD across a variety of model architectures, considering both similar and disparate student–teacher architectures. Besides, we constructed models to achieve varying compression levels.
We compared TSD's performance with that of a student trained from scratch (Base) and vanilla knowledge distillation (Base-KD). 
\Cref{tab:sim_arch} summarizes results on 28 datasets with teacher and student from the same model family, while \cref{tab:diff_arch} shows results for different model families.
Notably, TSD outperforms both the Base and Base-KD models by a large margin ($1.01\% \sim 10.66\%$). In every setting, TSD achieves the highest number of wins with the lowest average rank, establishing it as an efficient KD framework across different network architectures and compression levels.

\paragraph{TSD produces high-fidelity distillation.} 

While achieving good generalization, we further assess whether TSD also yields student models with higher fidelity. We distill an LSTM3-100 teacher into an LSTM1-8 student across the seven multi-class UCR datasets. \Cref{fig:fidelty_figsAll} presents the distillation fidelity, evaluated using both top-1 agreement (left) and predictive KL divergence (center) on the test set. Across all datasets, TSD achieves higher test agreement and lower predictive KL than vanilla KD.
In addition to evaluating fidelity on the test set, we also assessed it on the distillation dataset. The core intuition behind KD is that the student should, at a minimum, match the teacher on the distillation data in order to match it on the test set. 
However, prior work~\cite{stanton2021does} has revealed a fundamental trade-off between distillation and test fidelity.
We investigate how well TSD reacts to this trade-off in train-test fidelity using GAN-generated synthetic data~\cite{stanton2021does}. With the teacher trained on the original \textit{ItalyPowerDemand} dataset, we increase the dataset size by different factors during distillation. To eliminate any confounding from true labels, we only use distillation loss for the training. The corresponding train-test fidelity values are presented in \Cref{fig:fidelty_figsAll} (right). Results show that TSD outperforms vanilla KD, achieving higher train-test fidelity pairs across various distillation set expansion ratios.

\paragraph{TSD transfers interpretability.} 

To maintain the full functionality of the teacher, it is desirable for the student to preserve the same degree of interpretability. To assess transferred interpretability, we use occlusion analysis~\cite{zeiler2014visualizing}, a perturbation-based saliency evaluation method, along with three gradient-based saliency evaluation methods: GradSHAP~\cite{lindberg2017unified}, Integrated Gradients~\cite{sundararajan2017axiomatic}, and Gradient Saliency~\cite{simonyan2013deep}. We distill an LSTM3-100 teacher to an LSTM1-8 student across seven multi-class UCR datasets. \Cref{tab:explanation} reports the average MSE between teacher–student saliency-map pairs across all test samples. By achieving lower MSE values, TSD maintains a higher similarity with the teacher's saliency maps. This indicates that TSD tends to produce students that learn to use the same input features as the teacher models.
\begin{figure}[tb!]
    \centering
        \begin{minipage}[b]{0.23\textwidth}
        \centering
        \includegraphics[width=\textwidth]{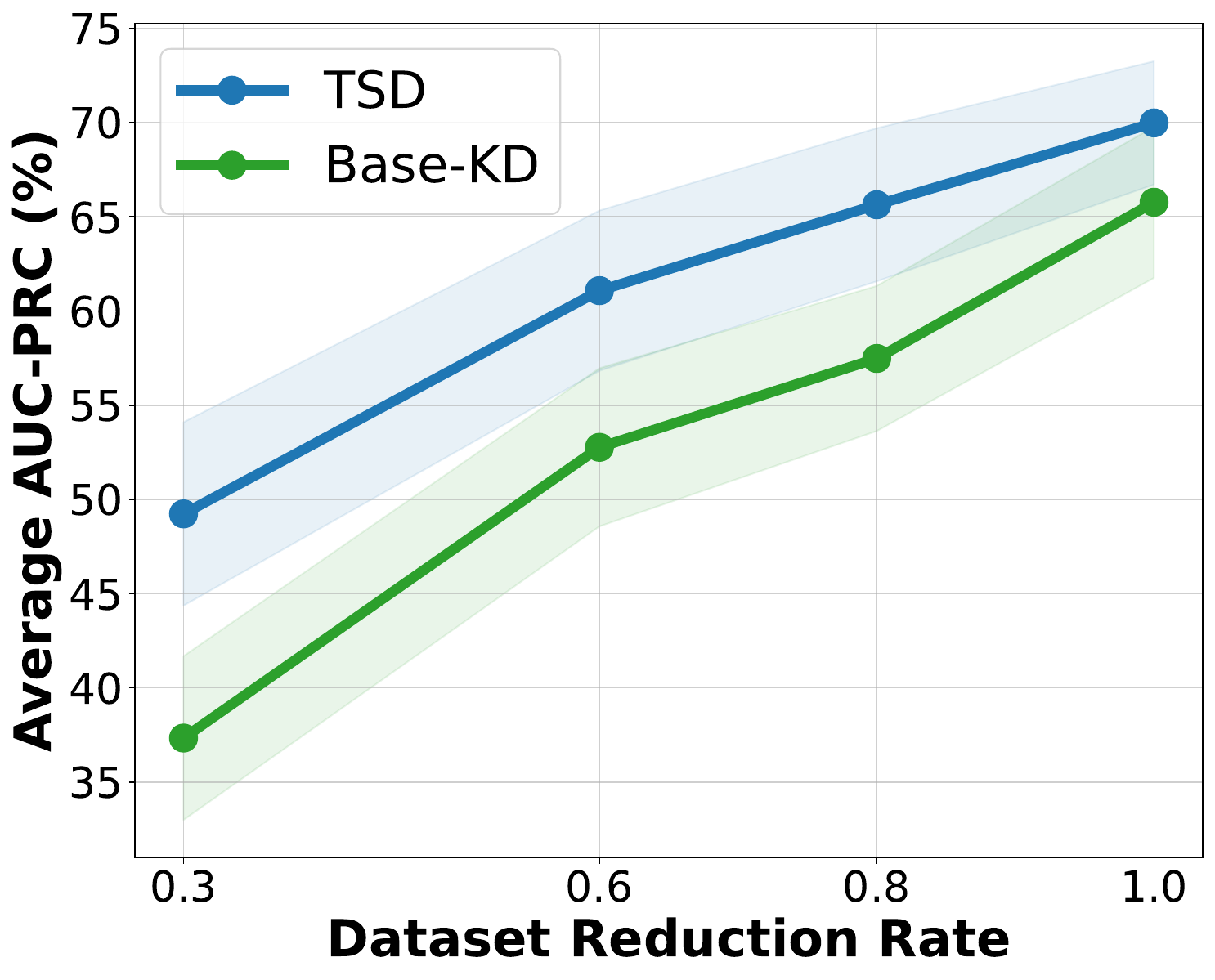}
    \end{minipage}
    \begin{minipage}[b]{0.23\textwidth}
        \centering
        \includegraphics[width=\textwidth]{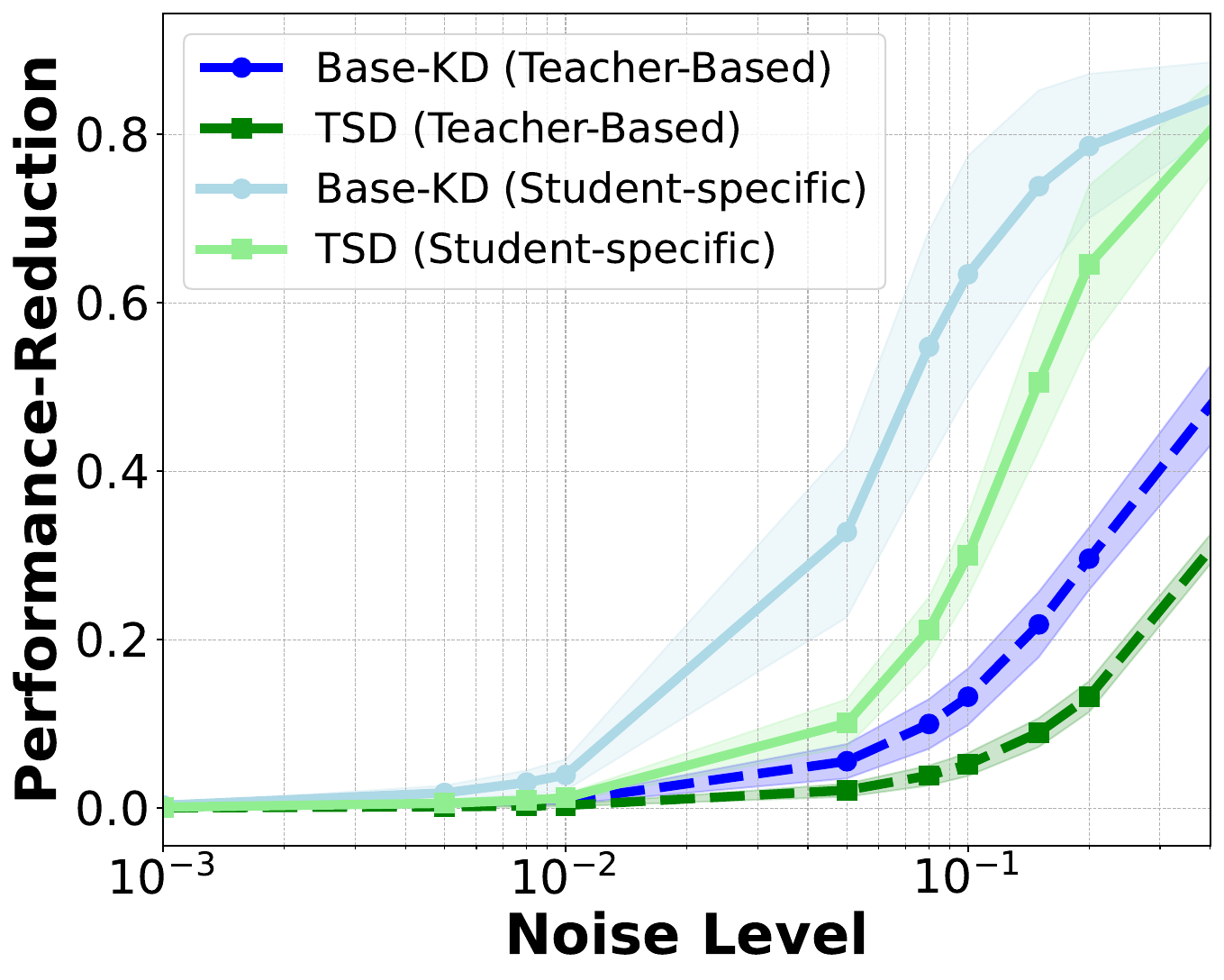}
    \end{minipage}
   \caption{\textbf{(Left)} Performance of students on seven multi-class UCR datasets where training set is reduced at various ratios. \textbf{(Right) }Performance (average AUC-PRC) degradation under varying adversarial noise levels. Both teacher-specific and student-specific FGSM noise are considered.}
    \label{fig:noise_robustness}
\end{figure}

\paragraph{TSD is robust to noise and limited data.} We compare the performance of TSD on seven multi-class UCR datasets, where the distillation set is reduced at various ratios, to evaluate its dependence on the amount of training data. We report the average AUC-PRC over datasets in \Cref{fig:noise_robustness} (left). The results show that TSD is minimally affected by the limitation of distillation samples compared to vanilla KD, i.e., TSD exhibits large gains with small distillation dataset sizes. Despite limited training samples, the TSD student sees more perturbed versions of the original samples during distillation, which may explain its robustness to limited data.
We further analyze TSD under FGSM noise, which generates adversarial perturbations, i.e., carefully crafted changes to the input, designed to deliberately fool a model into making incorrect predictions. We consider two scenarios: adversarial noise generated with respect to the 1) teacher and 2) each student. \Cref{fig:noise_robustness} (right) compares the performance (average AUC-PRC) degradation as the noise level increases at various ratios. In both scenarios, TSD's performance is minimally affected compared to vanilla KD.

\begin{table*}[tb!]
\caption{Similarity of saliency maps between the teacher and student on seven multi-class UCR datasets. Given the saliency map of the same input for the same prediction from both models, we used the mean squared (MSE) error to measure the similarity and report the average over the test set.}
\centering
\label{tab:explanation}
\begin{scriptsize}
\setlength{\tabcolsep}{6pt} 
\renewcommand{\arraystretch}{1.5} 
\begin{tabular}{c|cc|cc|cc|cc}
\textbf{Explanation } & \multicolumn{2}{c|}{Occlusion} & \multicolumn{2}{c|}{Grad shap} & \multicolumn{2}{c|}{Intregrated Grad.} & \multicolumn{2}{c}{Saliency} \\ 
\textbf{Dataset $\downarrow$}           & Base-KD   & TSD               & Base-KD   & TSD               & Base-KD             & TSD                 & Base-KD   & TSD              \\ 
UWaveGesture.All  & 0.0009    & \textbf{0.0005}   & 0.0976    & \textbf{0.0334}   & 0.2700              & \textbf{0.0094}     & 0.0033    & \textbf{0.0019}  \\
ShapesAll            & 0.0076    & \textbf{0.0046}   & 0.4242    & \textbf{0.1880}   & 13.4709             & \textbf{0.0949}     & 0.0246    & \textbf{0.0128}  \\
Adiac                & 0.0135    & \textbf{0.0074}   & 1.9175    & \textbf{1.3804}   & \textbf{3.2922}     & 3.5091              & 0.0352    & \textbf{0.0264}  \\
SwedishLeaf          & 0.0096    & \textbf{0.0025}   & 0.4138    & \textbf{0.1643}   & 0.4863              & \textbf{0.3087}     & 0.0298    & \textbf{0.0130}  \\
FaceAll              & 0.0056    & \textbf{0.0021}   & 0.1736    & \textbf{0.0683}   & 0.5587              & \textbf{0.1287}     & 0.0079    & \textbf{0.0027}  \\
NonInv.FatalECG2 & 0.0030    & \textbf{0.0013}   & 0.0594    & \textbf{0.0311}   & 1.3714              & \textbf{0.1299}     & 0.0067    & \textbf{0.0041}  \\
NonInv.FatalECG1 & 0.0032    & \textbf{0.0014}   & 0.0589    & \textbf{0.0314}   & 0.3009              & \textbf{0.1394}     & 0.0078    & \textbf{0.0032} \\ \bottomrule
\end{tabular}
\end{scriptsize}
\end{table*}

\begin{figure*}[tb!]
    \centering
    \begin{minipage}[b]{0.27\textwidth}
        \centering
        \includegraphics[width=\textwidth]{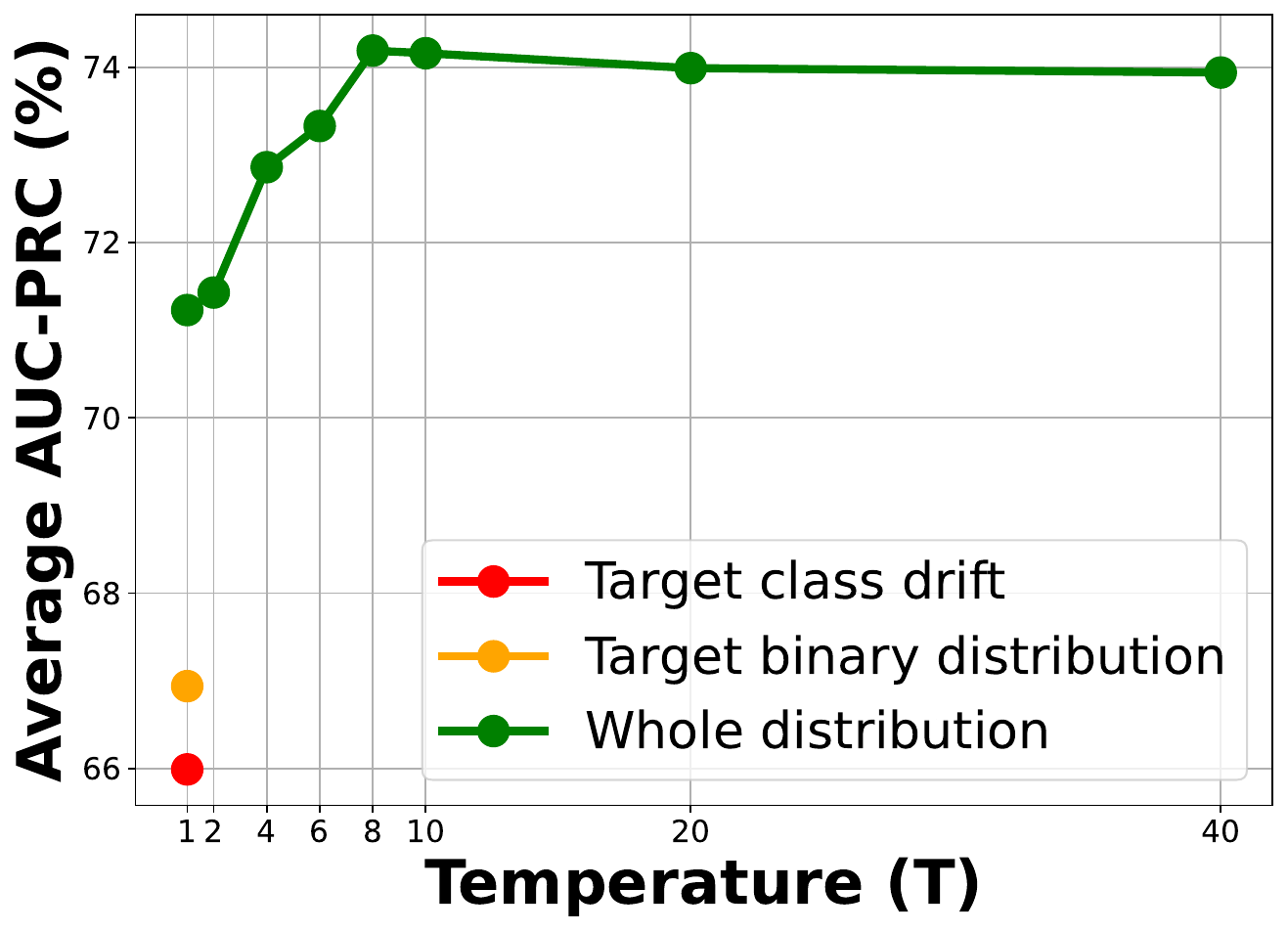}
    \end{minipage}
    \begin{minipage}[b]{0.33\textwidth}
        \centering
        \includegraphics[width=\textwidth]{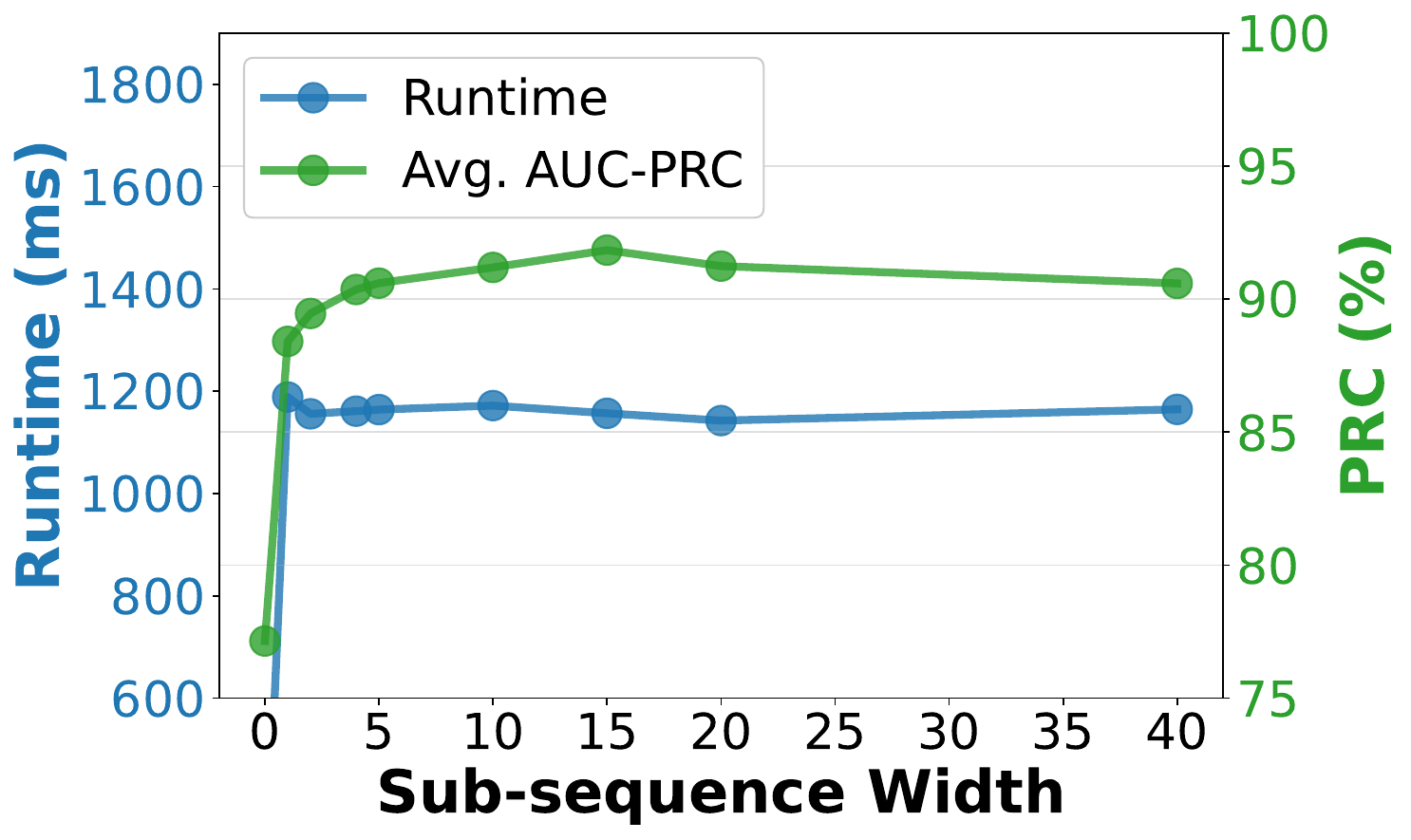}
    \end{minipage}
    \begin{minipage}[b]{0.33\textwidth}
        \centering
        \includegraphics[width=\textwidth]{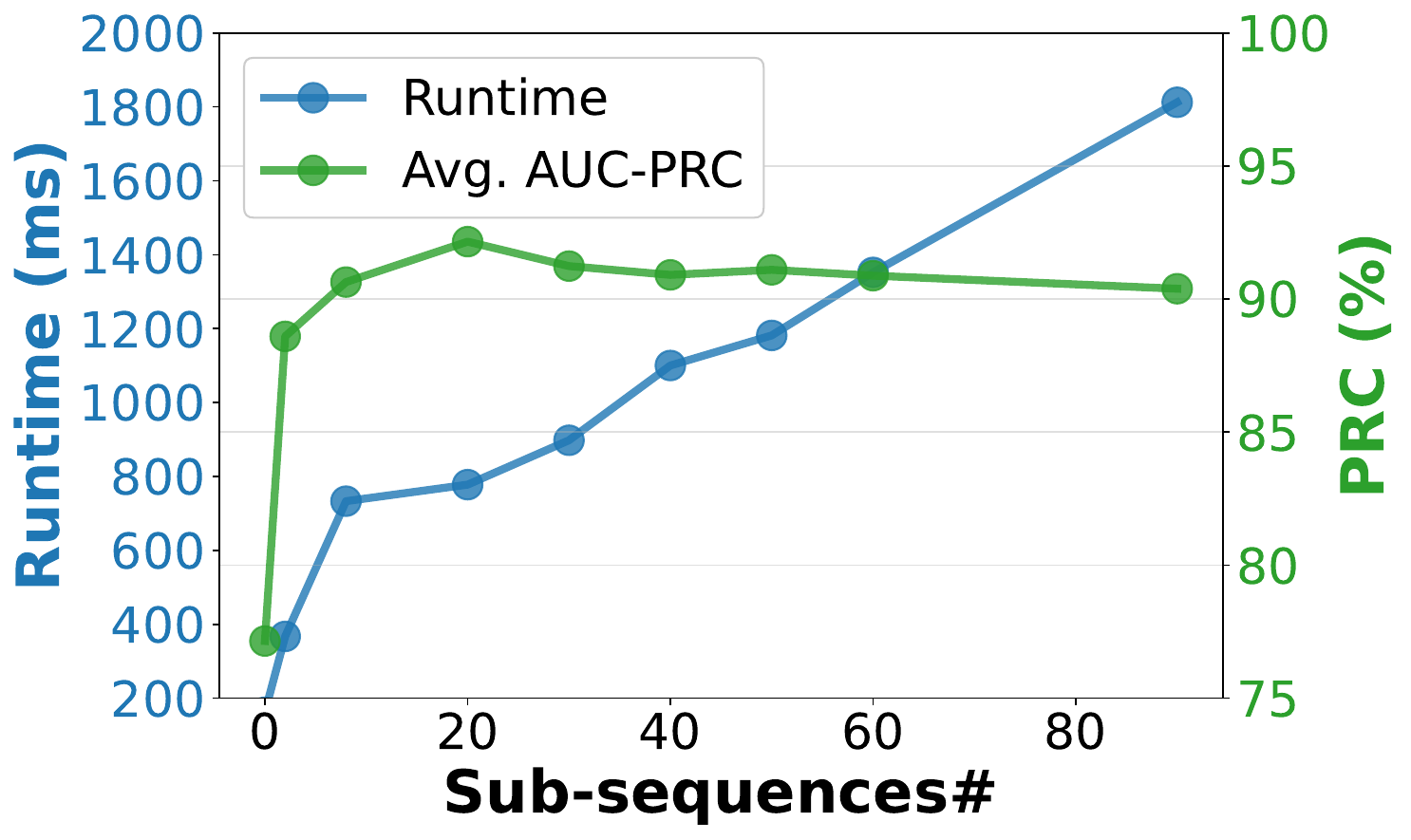}
    \end{minipage}
    \caption{\textbf{Analysis on TSD loss parameters}. (\textbf{Left)} Performance variations with the inclusion of non-target class details in the TSD loss. Three variants were evaluated to quantify the expected changes in model response: (1) scalar shift in the target class probability, (2) expected change in the binary class distribution, aggregating all non-target class effects (measured as KL divergence), and (3) expected change in the whole class distribution with individual effects of non-target classes (measured as KL divergence). Temperature scaling is used to further enhance non-target class details, with performance evaluated across different values of $\tau$. Performance is reported as the average AUC-PRC across seven UCR datasets, distilling an LSTM3-100 teacher into an LSTM1-8 student. \textbf{(Center)} Performance and runtime variations from ablation of the subsequence-width hyperparameter. \textbf{(Right)} Ablation study on the number of subsequnces.}
    \label{fig:ablation_params}
\end{figure*}

\subsection{Exploration of TSD}
\paragraph{Incorporating non-target class shifts enhances knowledge transfer when quantifying predictive distributional shifts. }\label{whole_effect}
There are a couple of ways we can measure how model outputs react to perturbations. A straightforward approach is to consider only the shift in the target class probability, as it directly reflects how the perturbation affects the model confidence in the target class. We are interested in understanding whether including changes in non-target classes in temporal saliency enhances knowledge transfer. We evaluate three forms of constructing temporal saliency: 1) quantifying the scalar shift in the target class probability: $|P(y_t\mid\boldsymbol{X}) - P(y_t \mid \tilde{\boldsymbol{X}})|$, where $P(y_t\mid.)$ represents target class probability predicted by the model, 2) quantifying the divergence between the original and perturbed outputs by modeling the output as a binary distribution (target class vs. all non-target classes)~\cite{zhao_decoupled_2022}: $\mathrm{KL}\left(P_\tau\left(y_t, y_{\backslash t} \mid \boldsymbol{X}\right) \| P_\tau\left(y_t,y_{\backslash t} \mid \tilde{\boldsymbol{X}}(t,z)\right)\right)$ and 3) quantifying the divergence across the entire output distribution, accounting for shifts in all classes: $\mathrm{KL}\left(P_\tau\left(\boldsymbol{Y} \mid \boldsymbol{X}\right) \| P_\tau\left(\boldsymbol{Y} \mid \tilde{\boldsymbol{X}}(t,z)\right)\right)$. \Cref{fig:ablation_params} (left) summarizes the performance of TSD variants across seven UCR datasets, where an LSTM3-100 teacher is distilled into an LSTM1-8 student.
The first metric (\textit{target class drift only}) results in the lowest performance. Including non-target class details in temporal saliency—using \textit{target binary distribution} and \textit{whole distribution}—significantly improves the performance of TSD. 
The \textit{whole distribution}, which captures individual changes in all non-target classes, significantly outperforms \textit{target binary distribution}, which aggregates changes across non-target classes. 
To further enhance non-target class details, we applied temperature scaling (see the next section).


\paragraph{When quantifying the predictive distributional shift, temperature scaling improves the effectiveness of knowledge transfer. }\label{T_effect}
We studied the impact of temperature scaling on TSD using different $\tau$ values, with results across seven UCR datasets summarized in \Cref{fig:ablation_params} (left). 
While $\tau=1$ recovers the original probability distribution, increasing $\tau$ raises the output entropy and elevates the importance of all class logits. Performance consistently improved with higher temperatures up to $\tau=8$, after which the gains diminished. This suggests that moderate temperature values enable TSD to better capture distributional changes, potentially revealing inter-class relationships, while excessive values may introduce unnecessary uncertainty and information loss.
Based on these findings, we set $\tau=8$ for all experiments.

\paragraph{Ablation study on subsequence width and number of subsequences.}\label{ablation_params}
\Cref{fig:ablation_params} (center) presents the ablation study on the subsequence width $z$, a key hyperparameter in the TSD loss function. The experiment distills an LSTM3-100 teacher to an LSTM2-32 student on the \textit{UWaveGestureLibraryAll} dataset. Student performance is reported for varying values of $z$, with the temperature fixed at 1 and 50 subsequences selected to evenly cover the entire time series. Performance increases with subsequence width up to $z=5$, beyond which it plateaus. We attribute this plateau to the fact that, with 50 subsequences, a width of $z \geq 5$ provides sufficient coverage for a time series of length 100. The runtime remained constant across different widths. Using the same experimental setup, \Cref{fig:ablation_params} (right) presents the ablation study on $\lvert\{t\}\rvert$: the cardinality of the set of starting indices of subsequences, which determines the number of selected subsequences used in the TSD loss. We fixed the temperature at 1 and set $z = 10$, where we observed that using 20 or more subsequences yields better performance. Although the runtime scales linearly with the number of subsequences, good performance can be achieved without resorting to higher values. For our experiments, we selected 50 subsequences with $z = 5$.

\section{Conclusion}
Knowledge Distillation is understudied in time series analysis, with no prior work utilizing knowledge about class-discriminative, salient time steps to improve the student model. In this paper, we propose Temporal Saliency Distillation that achieves both high interpretability and competitive performance. TSD incorporates a novel loss function that quantifies the importance of each time step to the teacher’s prediction through perturbation-based predictive distribution shifts and transfers this knowledge to the student model. TSD outperforms state-of-the-art KD methods and demonstrates robustness across different architectures, whether the student and teacher share the same or different architectures. We hope our findings encourage future research on enhancing the interpretability of KD.



\begin{ack}
This research was supported by The University of Melbourne’s Research Computing Services and the Petascale Campus Initiative. 
We also gratefully acknowledge the authors and maintainers of the UCR Time Series Archive~\cite{chen_ucr_2015} for making their datasets publicly available.
\end{ack}



\bibliography{refs}

@String(CVPR  = {IEEE Conf. Comput. Vis. Pattern Recog.})

@String(ICCV  = {Int. Conf. Comput. Vis.})

@String(NeurIPS = {Adv. Neural Inform. Process. Syst.})

@String(ICML  = {Int. Conf. Mach. Learn.})

@String(IJCAI = {IJCAI})

@String(ICASSP=	{ICASSP})

@String(IJCNN   = {International Joint Conference on Neural Networks (IJCNN)})

@String(CIKM   = {ACM International Conference on Information \& Knowledge Management})

@String(SIGKDD   = {ACM international conference on Knowledge discovery and data mining})

@String(ICDM   = {IEEE International Conference on Data Mining})

@String(ICTAI   = {IEEE conference on tools with artificial intelligence})

@String(BigData   = {EEE International Conference on Big Data})

@String(ICIM   = {International conference on information management})

@String(ICCBR   = {International conference on case-based reasoning})

@String(CVPR  = {CVPR})

@String(ICCV  = {ICCV})

@String(NeurIPS = {NeurIPS})

@String(ICML  = {ICML})

@String(IJCNN    = {IJCNN})

@String(CIKM    = {CIKM})

@String(SIGKDD    = {SIGKDD})

@String(ICDM    = {ICDM})

@String(ICTAI    = {ICTAI})

@String(BigData    = {BigData})

@String(ICIM    = {ICIM})

@String(ICCBR   = {ICCBR})

@article{ismail2020inceptiontime,
  title={Inceptiontime: Finding alexnet for time series classification},
  author={Ismail Fawaz, Hassan and Lucas, Benjamin and Forestier, Germain and Pelletier, Charlotte and Schmidt, Daniel F and Weber, Jonathan and Webb, Geoffrey I and Idoumghar, Lhassane and Muller, Pierre-Alain and Petitjean, Fran{\c{c}}ois},
  journal={Data Mining and Knowledge Discovery},
  volume={34},
  number={6},
  pages={1936--1962},
  year={2020},
  publisher={Springer}
}

@inproceedings{zeiler2014visualizing,
  title={Visualizing and understanding convolutional networks},
  author={Zeiler, Matthew D and Fergus, Rob},
  booktitle={Computer Vision--ECCV 2014: 13th European Conference, Zurich, Switzerland, September 6-12, 2014, Proceedings, Part I 13},
  pages={818--833},
  year={2014},
  organization={Springer}
}

@article{simonyan2013deep,
  title={Deep inside convolutional networks: Visualising image classification models and saliency maps},
  author={Simonyan, Karen and Vedaldi, Andrea and Zisserman, Andrew},
  journal={arXiv preprint arXiv:1312.6034},
  year={2013}
}

@inproceedings{sundararajan2017axiomatic,
  title={Axiomatic attribution for deep networks},
  author={Sundararajan, Mukund and Taly, Ankur and Yan, Qiqi},
  booktitle={International conference on machine learning},
  pages={3319--3328},
  year={2017},
  organization={PMLR}
}

@inproceedings{bucilua2006model,
  title={Model compression},
  author={Buciluǎ, Cristian and Caruana, Rich and Niculescu-Mizil, Alexandru},
  booktitle=SIGKDD,
  pages={535--541},
  year={2006}
}

@inproceedings{guilleme2019agnostic,
  title={Agnostic local explanation for time series classification},
  author={Guillem{\'e}, Ma{\"e}l and Masson, V{\'e}ronique and Roz{\'e}, Laurence and Termier, Alexandre},
  booktitle=ICTAI,
  pages={432--439},
  year={2019},
  organization={IEEE}
}

@article{vaswani2017attention,
  title={Attention is all you need},
  author={Vaswani, A},
  journal=NeurIPS,
  year={2017}
}

@article{tonekaboni2020went,
  title={What went wrong and when? Instance-wise feature importance for time-series black-box models},
  author={Tonekaboni, Sana and Joshi, Shalmali and Campbell, Kieran and Duvenaud, David K and Goldenberg, Anna},
  journal=NeurIPS,
  volume={33},
  pages={799--809},
  year={2020}
}

@inproceedings{lindberg2017unified,
  title={A unified approach to interpreting model prediction},
  author={Lindberg, SM and Lee, SI},
  booktitle={31st Conference on Neural Information Processing Systems (NIPS 2017)},
  year={2017}
}

@inproceedings{selvaraju2017grad,
  title={Grad-cam: Visual explanations from deep networks via gradient-based localization},
  author={Selvaraju, Ramprasaath R and Cogswell, Michael and Das, Abhishek and Vedantam, Ramakrishna and Parikh, Devi and Batra, Dhruv},
  booktitle=ICCV,
  pages={618--626},
  year={2017}
}

@inproceedings{zhou2016learning,
  title={Learning deep features for discriminative localization},
  author={Zhou, Bolei and Khosla, Aditya and Lapedriza, Agata and Oliva, Aude and Torralba, Antonio},
  booktitle=CVPR,
  pages={2921--2929},
  year={2016}
}

@article{stanton2021does,
  title={Does knowledge distillation really work?},
  author={Stanton, Samuel and Izmailov, Pavel and Kirichenko, Polina and Alemi, Alexander A and Wilson, Andrew G},
  journal=NeurIPS,
  volume={34},
  pages={6906--6919},
  year={2021}
}

@inproceedings{alharbi2021learning,
  title={Learning interpretation with explainable knowledge distillation},
  author={Alharbi, Raed and Vu, Minh N and Thai, My T},
  booktitle=BigData,
  pages={705--714},
  year={2021},
  organization={IEEE}
}

@inproceedings{zeyu2023grad,
  title={A grad-cam-based knowledge distillation method for the detection of tuberculosis},
  author={Zeyu, Ding and Yaakob, Razali and Azman, Azreen and Rum, Siti Nurulain Mohd and Zakaria, Norfadhlina and Nazri, Azree Shahril Ahmad},
  booktitle=ICIM,
  pages={72--77},
  year={2023},
  organization={IEEE}
}

@article{parchami2024good,
  title={Good Teachers Explain: Explanation-Enhanced Knowledge Distillation},
  author={Parchami-Araghi, Amin and B{\"o}hle, Moritz and Rao, Sukrut and Schiele, Bernt},
  journal={arXiv preprint arXiv:2402.03119},
  year={2024}
}

@inproceedings{fong2019understanding,
  title={Understanding deep networks via extremal perturbations and smooth masks},
  author={Fong, Ruth and Patrick, Mandela and Vedaldi, Andrea},
  booktitle=ICCV,
  pages={2950--2958},
  year={2019}
}

@inproceedings{crabbe2021explaining,
  title={Explaining time series predictions with dynamic masks},
  author={Crabb{\'e}, Jonathan and Van Der Schaar, Mihaela},
  booktitle=ICML,
  pages={2166--2177},
  year={2021},
  organization={PMLR}
}

@inproceedings{guo2023class,
  title={Class attention transfer based knowledge distillation},
  author={Guo, Ziyao and Yan, Haonan and Li, Hui and Lin, Xiaodong},
  booktitle=CVPR,
  pages={11868--11877},
  year={2023}
}

@inproceedings{delaney2021instance,
  title={Instance-based counterfactual explanations for time series classification},
  author={Delaney, Eoin and Greene, Derek and Keane, Mark T},
  booktitle=ICCBR,
  pages={32--47},
  year={2021},
  organization={Springer}
}

@article{briandet1996discrimination,
  title={Discrimination of Arabica and Robusta in instant coffee by Fourier transform infrared spectroscopy and chemometrics},
  author={Briandet, Romain and Kemsley, E Katherine and Wilson, Reginald H},
  journal={Journal of agricultural and food chemistry},
  volume={44},
  number={1},
  pages={170--174},
  year={1996},
  publisher={ACS Publications}
}

@article{paszke_2019_pytorch,
  title={Pytorch: An imperative style, high-performance deep learning library},
  author={Paszke, Adam and Gross, Sam and Massa, Francisco and Lerer, Adam and Bradbury, James and Chanan, Gregory and Killeen, Trevor and Lin, Zeming and Gimelshein, Natalia and Antiga, Luca and others},
  journal=NeurIPS,
  volume={32},
  year={2019}
}

@inproceedings{wang2017time,
  title={Time series classification from scratch with deep neural networks: A strong baseline},
  author={Wang, Zhiguang and Yan, Weizhong and Oates, Tim},
  booktitle=IJCNN,
  pages={1578--1585},
  year={2017},
  organization={IEEE}
}

@article{ba_deep_2014,
  title={Do deep nets really need to be deep?},
  author={Ba, Jimmy and Caruana, Rich},
  journal=NeurIPS,
  volume={27},
  year={2014}
}

@article{hinton_distilling_2014,
  title={Distilling the knowledge in a neural network},
  author={Hinton, Geoffrey and Vinyals, Oriol and Dean, Jeff},
  journal={arXiv preprint arXiv:1503.02531},
  year={2015}
}

@misc{romero_fitnets_2014,
      title={FitNets: Hints for Thin Deep Nets}, 
      author={Adriana Romero and Nicolas Ballas and Samira Ebrahimi Kahou and Antoine Chassang and Carlo Gatta and Yoshua Bengio},
      year={2015},
      eprint={1412.6550},
      archivePrefix={arXiv},
      primaryClass={id='cs.LG' full_name='Machine Learning' is_active=True alt_name=None in_archive='cs' is_general=False description='Papers on all aspects of machine learning research (supervised, unsupervised, reinforcement learning, bandit problems, and so on) including also robustness, explanation, fairness, and methodology. cs.LG is also an appropriate primary category for applications of machine learning methods.'}
}

@inproceedings{parvatharaju_learning_2021,
  title={Learning saliency maps to explain deep time series classifiers},
  author={Parvatharaju, Prathyush S and Doddaiah, Ramesh and Hartvigsen, Thomas and Rundensteiner, Elke A},
  booktitle=CIKM,
  pages={1406--1415},
  year={2021}
}

@inproceedings{ye_time_2009,
  title={Time series shapelets: a new primitive for data mining},
  author={Ye, Lexiang and Keogh, Eamonn},
  booktitle=SIGKDD,
  pages={947--956},
  year={2009}
}

@inproceedings{patel_mining_2002,
  title={Mining motifs in massive time series databases},
  author={Patel, Pranav and Keogh, Eamonn and Lin, Jessica and Lonardi, Stefano},
  booktitle=ICDM,
  pages={370--377},
  year={2002},
  organization={IEEE}
}

@inproceedings{zhu_time_2018,
  title={Time Series Chains: A Novel Tool for Time Series Data Mining.},
  author={Zhu, Yan and Imamura, Makoto and Nikovski, Daniel and Keogh, Eamonn J},
  booktitle=IJCAI,
  pages={5414--5418},
  year={2018}
}

@article{zagoruyko_paying_2017,
  title={Paying more attention to attention: Improving the performance of convolutional neural networks via attention transfer},
  author={Zagoruyko, Sergey and Komodakis, Nikos},
  journal={arXiv preprint arXiv:1612.03928},
  year={2016}
}

@inproceedings{park_relational_2019,
  title={Relational knowledge distillation},
  author={Park, Wonpyo and Kim, Dongju and Lu, Yan and Cho, Minsu},
  booktitle=CVPR,
  pages={3967--3976},
  year={2019}
}

@article{chen_ucr_2015,
  title={The UCR time series archive},
  author={Dau, Hoang Anh and Bagnall, Anthony and Kamgar, Kaveh and Yeh, Chin-Chia Michael and Zhu, Yan and Gharghabi, Shaghayegh and Ratanamahatana, Chotirat Ann and Keogh, Eamonn},
  journal={IEEE/CAA Journal of Automatica Sinica},
  volume={6},
  number={6},
  pages={1293--1305},
  year={2019},
  publisher={IEEE}
}

@inproceedings{ahn_variational_2019,
  title={Variational information distillation for knowledge transfer},
  author={Ahn, Sungsoo and Hu, Shell Xu and Damianou, Andreas and Lawrence, Neil D and Dai, Zhenwen},
  booktitle=CVPR,
  pages={9163--9171},
  year={2019}
}

@inproceedings{zhao_decoupled_2022,
  title={Decoupled knowledge distillation},
  author={Zhao, Borui and Cui, Quan and Song, Renjie and Qiu, Yiyu and Liang, Jiajun},
  booktitle=CVPR,
  pages={11953--11962},
  year={2022}
}

@inproceedings{qiao_class-incremental_2023,
  title={Class-Incremental Learning on Multivariate Time Series Via Shape-Aligned Temporal Distillation},
  author={Qiao, Zhongzheng and Hu, Minghui and Jiang, Xudong and Suganthan, Ponnuthurai Nagaratnam and Savitha, Ramasamy},
  booktitle=ICASSP,
  pages={1--5},
  year={2023},
  organization={IEEE}
}

@article{xing_2022_efficient,
  title={An efficient federated distillation learning system for multitask time series classification},
  author={Xing, Huanlai and Xiao, Zhiwen and Qu, Rong and Zhu, Zonghai and Zhao, Bowen},
  journal={IEEE Transactions on Instrumentation and Measurement},
  volume={71},
  pages={1--12},
  year={2022},
  publisher={IEEE}
}

@article{campos_2023_lightts,
  title={LightTS: Lightweight time series classification with adaptive ensemble distillation},
  author={Campos, David and Zhang, Miao and Yang, Bin and Kieu, Tung and Guo, Chenjuan and Jensen, Christian S},
  journal={Proceedings of the ACM on Management of Data},
  volume={1},
  number={2},
  pages={1--27},
  year={2023},
  publisher={ACM New York, NY, USA}
}

@article{hochreiter1997long,
  title={Long Short-term Memory},
  author={Hochreiter, S},
  journal={Neural Computation MIT-Press},
  year={1997}
}

@misc{hewa_dehigahawattage_2025_16938636,
  author       = {Hewa Dehigahawattage, Nilushika Udayangani},
  title        = {Supplementary Information for "Learning to Reason:
                   Temporal Saliency Distillation for Interpretable
                   Knowledge Transfer"
                  },
  month        = aug,
  year         = 2025,
  publisher    = {Zenodo},
  doi          = {10.5281/zenodo.16938636},
  url          = {https://doi.org/10.5281/zenodo.16938636},
}

\end{document}